\documentclass[final,3p,times]{elsarticle}
\usepackage{amsmath,amssymb,amsfonts}
\usepackage{amsthm}
\usepackage{graphicx}
\usepackage{hyperref}
\usepackage{algc}
\usepackage{xcolor}

\definecolor{mycolor}{rgb}{0.7,0.3,0.3}

\newcommand{\Real}{\mathbb{R}}
\newcommand{\bfx}{\boldsymbol{x}}

\newcommand{\bfw}{\boldsymbol{w}}

\usepackage{lineno}

\usepackage{amssymb,amsmath,graphicx}
\usepackage{epigraph}

\def\tr{{\raise0pt\hbox{$\scriptscriptstyle\top$}}}

\newtheorem{example}{Example}
\newtheorem{proposition}{Proposition}
\newtheorem{theorem}{Theorem}
\newtheorem{remark}{Remark}
\newtheorem{corollary}{Corollary}
\newtheorem{definition}{Definition}

\begin{document}

\begin{frontmatter}

\title{The unreasonable effectiveness of small neural ensembles \\ in high-dimensional brain}

\author[LeicMath,NNU]{A.N. Gorban\corref{cor1}}
\ead{a.n.gorban@le.ac.uk}
\author[NNU,UCM]{V.A. Makarov}
\ead{vmakarov@ucm.es}
\author[LeicMath,NNU,LETI]{I.Y. Tyukin}
\ead{i.tyukin@le.ac.uk}

\address[LeicMath]{Department of Mathematics, University of Leicester, Leicester, LE1 7RH, UK}
\address[NNU]{Lobachevsky University, Nizhni Novgorod, Russia}
\address[UCM]{Instituto de Matem\'{a}tica Interdisciplinar, Faculty of Mathematics, Universidad Complutense de Madrid, Avda Complutense s/n, 28040 Madrid, Spain}
\address[LETI]{Saint-Petersburg State Electrotechnical University, Saint-Petersburg,  Russia}

\cortext[cor1]{Corresponding author}

\begin{abstract}

Complexity is an indisputable, well-known,  and broadly accepted feature of the brain. Despite the {apparently} obvious and widely-spread consensus on the brain complexity, {sprouts of}  the {\em single neuron revolution} emerged in neuroscience in  the 1970s. {They brought} many unexpected discoveries, including grandmother or concept cells and sparse coding of information in the brain.

In machine learning for a long time, the famous curse of dimensionality seemed to be an unsolvable problem. Nevertheless, the idea of the {\em blessing of dimensionality} becomes gradually more and more popular. {Ensembles} of non-interacting or weakly interacting simple units prove to be an effective tool for solving essentially multidimensional  {and apparently incomprehensible} problems. This approach is especially useful {for} one-shot (non-iterative) correction of errors {in} large legacy artificial intelligence systems {and} when the complete re-training is impossible or too expensive.

These simplicity revolutions {in} the era of complexity have deep  {fundamental reasons  grounded} in geometry of multidimensional data spaces. To explore and understand these  reasons we   {revisit} the background ideas of statistical physics.  In the course of the 20th century they were developed  into  {the} concentration of measure theory. 
The Gibbs equivalence of ensembles with further generalizations shows that the data in high-dimensional spaces are concentrated near shells of smaller dimension.  New stochastic separation theorems reveal the fine structure of the data clouds.

We review and analyse biological, physical, and mathematical problems   at the core of the fundamental question: how can high-dimensional brain  organise reliable and fast learning in high-dimensional world of data by simple tools? To meet this challenge,  we outline and setup a framework based on statistical physics of data.

Two critical applications are reviewed to exemplify the approach: one-shot correction of errors in intellectual systems and emergence  of static and associative memories in ensembles of single neurons. Error  correctors should  be simple; not damage the existing  skills of the system; allow fast non-iterative learning  and  correction of  new mistakes without destroying the previous fixes.  All these demands can be satisfied by new tools based on the concentration of measure phenomena and stochastic separation theory. 

We show how a simple enough functional neuronal model is capable of explaining: i) the extreme selectivity of single neurons to the information content of high-dimensional data, ii) simultaneous separation of several uncorrelated   informational items from a large set of stimuli, and iii) dynamic learning of new items by associating them with already “known” ones. These results constitute a basis for organisation of complex memories in ensembles of single neurons.

\end{abstract}


\end{frontmatter}

\epigraph{Nature, just like the Sphinx, contrives to set \\
Mankind the deadliest test that ever was. \\
Why do we always fail? Perhaps because \\
She holds no riddle, and has never yet. }{Fyodor Tyutchev, Ovstug. August 1869, \\Translated by John Dewey \cite{Tyutchev}}

\section{Introduction}

Physics aims at describing  the widest   {realms} of reality   {with} as few   {basic or fundamental} principles as possible.   {Over centuries}, from Newton to Einstein, the main   {strive was to  discover} the `laws of Nature'. These laws have to be beautiful and simple, and the explained portion of reality should be as large as possible. Nevertheless, according to Feynman, `We never are definitely right, we can only be sure we are wrong. ...  In other words we are trying to prove ourselves wrong as quickly as possible, because only in that way can we find progress' \cite{FeynmanCharacter}.

\begin{figure}[h]
\centering{
\includegraphics[width=0.9 \textwidth]{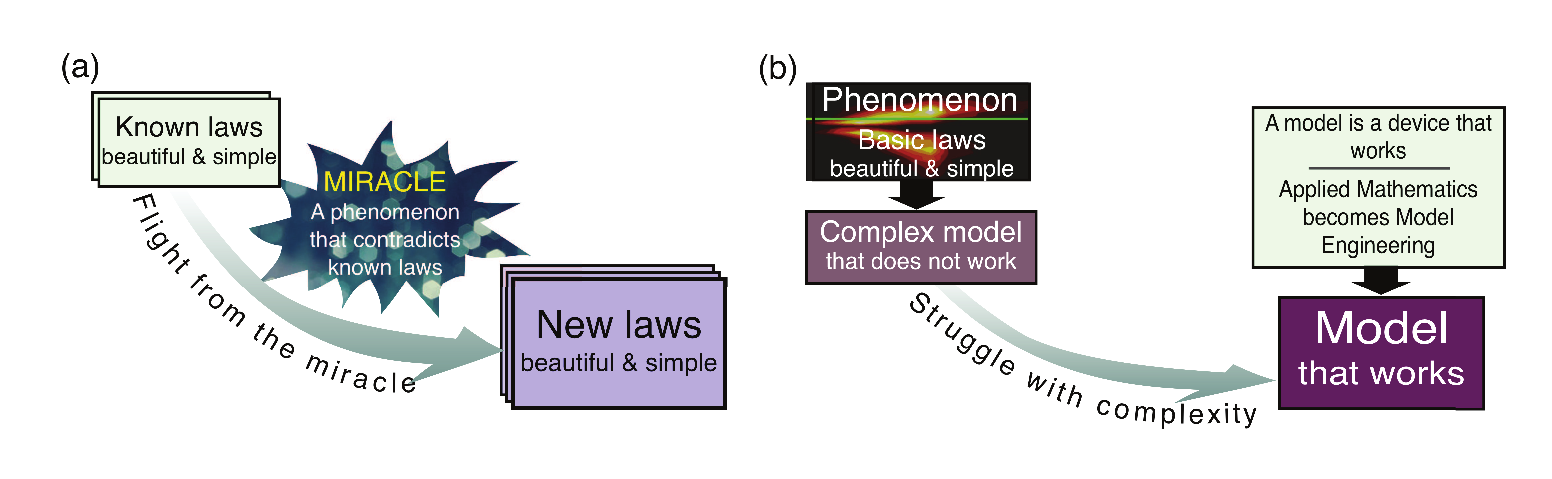} 
\caption{\label{Fig:Flight}(a) The flight from miracle: Einstein's road; (b) The struggle with complexity: the life battle  of the model engineers.}}
\end{figure}

Once we  proved our   {current} laws wrong,   {new} laws are needed.    {Discovering new laws, according to Einstein, can be viewed as} a `flight from miracle' \cite{Einstein}: `The development
of this world of thought is in a certain sense  a continuous flight from `miracle'.' What
does it mean? Let us imagine: we have   laws, beautiful and simple (the Newtonian
mechanics, for example). Then  we find a phenomenon that   {these laws} cannot   {explain}.
This is a {\em miracle}, a phenomenon that contradicts the basis laws. However, we like these laws
and try to use them again and again to describe the miracle. If we fail then we have to
use another way. We like our laws but we   {like} rationality more,   {and} therefore we fly   {away} from
the miracle by inventing new laws, which are beautiful, simple and, at the same time,
  {explain} the phenomenon. After that, the miracle disappears and we have new
laws, beautiful and simple (Fig.~\ref{Fig:Flight} a) \cite{GorbanMainstream2006}.

  {The cycle of creation of new laws, their analysis followed by deduction of their consequences and drawing predictions requires an effective machinery. Mathematics has long been recognized as such.} Wigner found  this effectiveness mysterious and unreasonable \cite{Wigner1960}: `The miracle of the appropriateness of the language of mathematics for the formulation of the laws of physics is a wonderful gift which we neither understand nor deserve.' There were many attempts to explain this miracle. The main idea was: Mathematics develops as a gradual ascent on stairs of abstractions. It consists of tools to solve problems, that  have arisen in the development of tools to solve problems, that have arisen..., ... in developing tools to solve real life problems. It   {is hence not surprising} that such a tool for tools, a beautiful metatool or megatool,   {is indeed} effective downstairs.

The face of science is changing gradually. The phrase of the great physicist Stephen Hawking, `I think the next century will be the century of complexity,' in his `millennium' interview on January 23, 2000 (San Jose Mercury News) became a widely cited prophecy. The idea  of `complexity'   {requires perhaps} even more   {elaborated clarification} than Einstein's `flight from miracle'. Large nonlinear systems and `emerging' phenomena have been studied long time ago. More than two thousand years ago Aristotle wrote that
`the whole is something besides the parts' (Metaphysics, Book 8, Chapter 6).   {Since then and to date, complexity traditionally is attributed to some objects and phenomena that are being studied. However, as has been advocated in \cite{GorbanYabl2013}, to truly understand what is it that constitutes complexity, instead of working with an object or phenomenon itself, analysing human behavior and activities (i.e. the struggle with complexity) arising whilst studying the phenomenon could sometimes be more productive. This leads to a simple yet fundamental conclusion that complexity is the gap betuween the laws and the phenomena \cite{GorbanYabl2013}.}

We can imagine a `detailed' model for a phenomenon but due to complexity, we
cannot work with this detailed model. For example, we can write the Schr\"odinger
equation for nuclei and electrons,  but we
cannot use it directly for modelling  materials or large molecules.
The result of the struggle with complexity is a model that works. This is   {reminiscent} of
{\em engineering}: a model is a device, and this device   {must} be functional.   {Different models are needed for different purposes}.   {One} may combine the first principles,  empirical data, and even   {conduct dedicated} active experiments to create   {suitably detailed} models. These imaginary detailed models are   {then} combined   {into}  `possible worlds'   {satisfying} the   {first-principled} laws completely. A crucial question arises: is an observed phenomenon possible in such a possible world?   

{In the context of mind and brain, examples of such possible worlds as well as the corresponding open questions are well-known (see review \cite{Perlovsky2006}). The variety of proposed answers is impressive. Minsky asked \cite{Minsky1988}: `What magical trick makes us intelligent?' and answered: `The trick is that there is no trick' -- no new principles are necessary.  Quite opposite, Penrose stated that the known physical principles are insufficient for the explanation of intelligence and mind \cite{Penrose1994}. But even for somewhat simpler fluid dynamics models in atomistic world the situation is not fully clear yet and  famous Hilbert's problem `of developing mathematically the limiting processes, ... which lead from the atomistic view to the laws of motion of continua' is not solved, moreover, the possible solution may be negative  \cite{Slemrod2018}.} These problems form an essential part of the 6th Hilbert's problem \cite{GorbanHilbert2018}.

Many toolboxes have been developed for model reduction, especially in such areas as physical and chemical kinetics, where the variety of models is huge \cite{TomlinEtAl1997,GorbanModRed2018}. Modern mathematical modelling is impossible without model reduction, and the model should be simplified during development \cite{GorbanModRed2018,GorbanKevrekidisMiltiscale2006}.

   {Model} reduction is necessary in the analysis of both natural and artificial neural networks   {too}.   {For example, dimensionality reduction is imperative} for defining the `genuine dimension' of the processed signals and images and for development lower-dimensional models. But surprisingly, simplicity of some basic models in neuroscience  is not  a result of model reduction. It does not come from    fundamental laws   {either}. It does not follow the fundamental schemes from Figs.~\ref{Fig:Flight} a,b.  We   consider this simplicity as a riddle. Further in this review we detail this riddle, step by step, and discuss its possible solution.

In neuroscience,   {like in physics}, some simple   {models} work   {reasonably} well.   {Recall, for instance,}  classical   {Pavlovian} conditioning   {and Hebb's} rule. But as we go deeper into   {details of} neuronal mechanisms of perception and memory, we expect complexity of models to increase, and the structural-functional relationships between brain structures and observed phenomena to become more sophisticated. Surprisingly, simple models of neuronal mechanisms of perception and memory demonstrate   {remarkable effectiveness}. 

\begin{figure}[h]
\centering{
\includegraphics[width=0.8\textwidth]{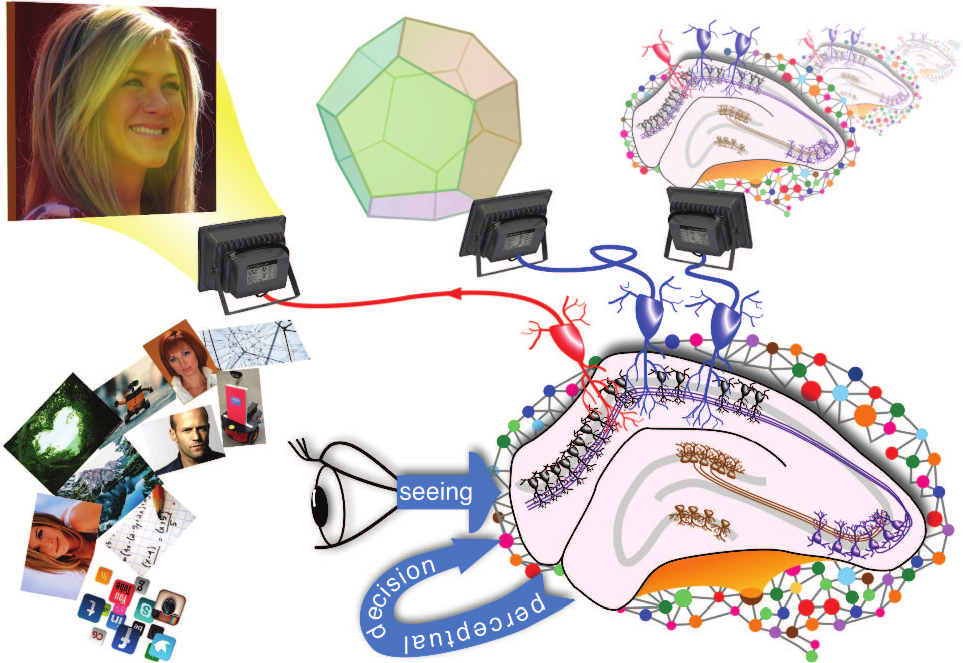} 
\caption{\label{Fig:Grandmother}Idea of grandmother cell, a neuron that reacts selectively on a pattern: Jennifer Aniston cell, Dodecahedron cell, and `Grandmother cell' cell.}}
\end{figure}

One basic example, {\em grandmother cells}, seemed to be so simplistic that the term was introduced by Lettvin around 1969 in a jocular story about `a great if unknown neurosurgeon', Akakhi Akakhievitch, who deleted concepts from patient's memory by ablating  the corresponding cells \cite{Gross2002}. More seriously, this is a hypothesis that there are neurons that react  selectively to specific concepts and images: there are grandmother cells, Jennifer Aniston cells, Dodecahedron cells of even `Grandmother cell' cells (the cells,  that react selectively on the pattern or  even the idea of grandmother cell) (see Fig.~\ref{Fig:Grandmother}).  Two years before Lettvin's joke, the very similar concept of {\em gnostic cells} was introduced by Konorski  in his  research monograph \cite{Konorski1967}.

There are several attempts to give a more  formal definition  of grandmother (or concept) cell. These definitions differ in their relation to real brain and these differences stimulate intensive discussions \cite{Bowers2009,Bowers2010,Plaut2010,QuianQuirogaKreiman2010,QuianQuirogaKreiman2010a}.

\begin{figure}[h]
\centering{
\includegraphics[width=0.9 \textwidth]{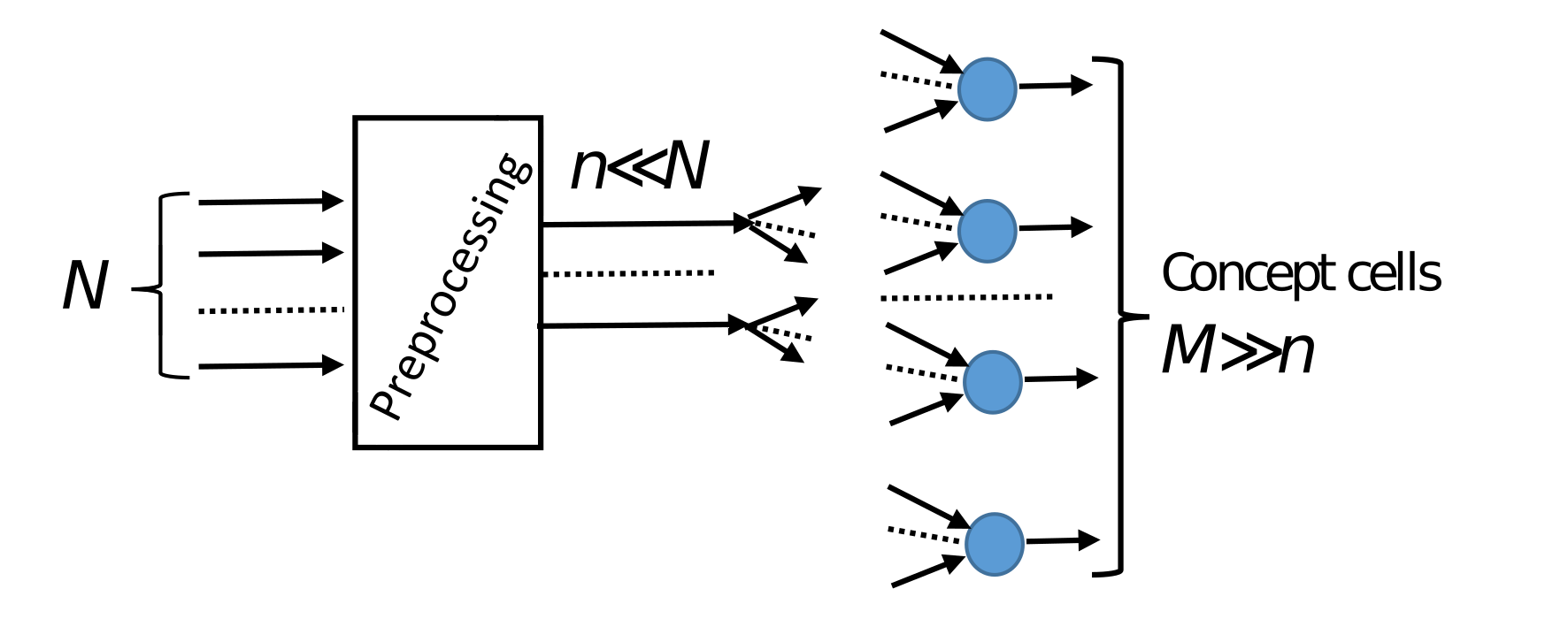} 
\caption{\label{Fig:Concept}Schematic representation of the idea behind idealised `concept cells': high-dimensional input signal  (dimension $N$)  after preprocessing (dimension $n$), arrives to an ensemble of non-interacting concept cells. The number of concept cells $M$ is much larger than $n$ (and may be even larger than $N$). Each concept cell reacts to a single concept in various presentations: visual, verbal, or even on imagination. A source of sparsity could be due to restrictions on the number of inputs of a single concept cell and on the number of links from one component of the preprocessed signal to different concept cells.}}
\end{figure}

The idea about one-to-one correspondence `one concept -- one cell' told in Lettvin's jocular story about  Akakhi Akakhievitch and grandmother cell is hard (if at all possible) to verify in an experiment. And yet, the main hypothetical properties of the idealised concept cells are not radically unusual if their broad interpretations are allowed (Fig.~\ref{Fig:Concept}):
\begin{itemize}
\item Selectivity -- the cell reacts to a single specific concept.  Of course, it is impossible to exhaustively test all concepts, and precise meaning and interpretation of that constitutes the cell's reaction is to be defined.  Moreover, association of concepts is also possible, and the formal notion of a single concept in brain study is not strictly defined too.  (Nevertheless, psychologists and brain scientists successfully work with this concept of a concept).
\item Invariance -- the cell reacts to various representations of the same  concept, including different visual images, verbal representation and even imagination.
\item Different concept cells receive components of the same vector of preprocessed signal.
\item The number of different concepts can be much larger than the dimension of the preprocessed signal (and even larger than the dimension of the inputs).
\end{itemize}

The concept of a grandmother cell   {did not emerge as an outcome of any model reduction procedure applied to a} detailed model. The brain modelling does not follow the model engineering `struggle with complexity' scheme (Fig.~\ref{Fig:Flight} b): there is no  complex model that follows the basic laws but does not work. On the contrary, the grandmother  cell is a simple   {phenomenological} construction that has been extracted from the discussion of the empirical world and used as a basic brick in modelling. Everyone understands the limitations of this simple model  and further development adds more specific details.

This methodology seems to be different from modern physics:
\begin{itemize}
\item In physics, we can imagine a `detailed model' that follows the basic laws. We consider a `possible world', that is the Universe, which follows the basic laws in full details. The notion of possible worlds \cite{Hintikka1967} was introduced by Leibnitz for semantics of science.  Of course, even here we do not have a complete self-consistent system of laws, but there is a web of standard models \cite{GorbanHilbert2018} with some contradictions that could be more of less hidden. Therefore, strictly speaking, even in physics we cannot speak of possible worlds, but rather of `impossible possible worlds' \cite{Hintikka1975,GorbanHilbert2018}. However, contradictions, such as the quantum gravity problem, seem to be far from most real life problems, and a possible world of physics can be presented along with the ideal `detailed model' for further simplification.
\item In neuroscience, a detailed physical model of the brain is not conceivable if we want it to be the model of the {\em brain} and do not like to consider the model of trivial generality such as some unknown ensemble of molecules, reactions and flows. Instead of a detailed model of the brain, a collection of elementary components like the grandmother cell are used, and the system of models is created. This system is developed further in various directions: to the upper levels, with the necessary specification and  addition of new details to describe and explain the integrative activity of the brain, and to more physical and technical levels, to decipher the physical and chemical mechanisms of elementary processes in the brain.
\end{itemize}

Simple models are at the centre of the developing system of models, and many of these simple models are grandmother cells or very simple neural ensembles with a certain semantic function. These simple models were not produced by reduction of more detailed models. They also do not represent the basic fundamental laws. This {\em simplicity without simplification} was guessed from observations and scientific speculations about the brain functioning and subsequently confirmed by various experiments that required at the same time modifications of   original models.  Such an approach, without systematic appellation to  basic laws and detailed models is also widely used in physics when the basic laws are largely unknown (see the classification of models in physics proposed by Peierls   \cite{Peierls1980}).

The idea that `each concept -- each person or thing in our everyday experience -- may have a set of
corresponding neurons assigned to it' was supported by  experimental evidence  and became very popular \cite{QuianQuirogaSciAm2013}. Of course, the `pure grandmother cells' are considered now as the ultimate abstraction. If two concepts appear often together, then the cell will react   {to} both. Moreover, this association could be remembered from a single exposure \cite{IsonQQ2015}. For example, the neuroscientists will associate  Jennifer Aniston with the concept of grandmother cell after the famous series of experiments, which used  pictures of this actress \cite{QuianQuirogaNature2005}. If somebody will see Jennifer Aniston with Dodecahedron in her hands then he (or she) will associate these images for a long time, and the corresponding Jennifer Aniston cell could be, at the same time, the Dodecahedron cell. Interaction between concepts can be also negative, for example,  it was reported that in some experiments the Jennifer Aniston unit did not respond to pictures of Jennifer Aniston together with   Brad Pitt: one image can outshine the other in perception \cite{QuianQuirogaNature2005}.

Deciphering of the concepts hidden in the activity of populations of neurons is a fascinating task. It requires combination of statistical methods and information theory  \cite{PougetEtAl2000,QuianQuirogPanzeri2009}.

The further complications can not overshadow the simplicity of the main construction: the idea of grandmother cell is unexpectedly efficient. Moreover, it is very close to experiment because the study of single-cell responses to various stimuli or behaviours is the basis of experimental neuroscience despite  the obvious expectations that the brain   makes decisions by processing the activity of large
neuronal populations \cite{QuianQuirogPanzeri2009}. After decades of development of the initial idea, the neural coding of concepts is considered as sparse but it is not a single-cell  coding \cite{QuianQuirogaTrends2008}.

The brain processes intensive flow of high-dimensional information. For example, simulation of the realistic-scale brain models \cite{IzhikevichEdelman2008} reported in 2005 involved $10^{11}$ neurons and almost $10^{15}$ synapses. One second of simulation took 50 days on a Beowulf cluster of 27 processors, 3GHz each.

A natural question arises: is there a fundamental reason for the emergence of single  grandmother cells or small neuronal ensembles with a certain semantic function in such a multidimensional brain system and information flow? We aim to answer this question and to demonstrate that the answer is likely to be `yes': There are very deep reasons for the appearance of such elementary structures in a high-dimensional brain. This is a manifestation of a general {\em blessing of dimensionality} phenomenon.

The blessing of dimensionality is based on the theory of measure concentration phenomena \cite {Bal1997,GianMilman2000,Gromov2003,Ledoux2005,Talagrand1995} and the stochastic separation theorems  \cite{GorbanGrechukTykin2018,GorbTyu2017,GorbanTyuRom2016}. These results form the basis of the  `blessing of dimensionality' in machine learning \cite {AndersonEtAl2014,Donoho2000,DonohoTanner2009,GorbanGrechukTykin2018,GorbanTyuRom2016,Kainen1997}.

\begin{figure}
\begin{centering}
\includegraphics[width=0.9 \textwidth]{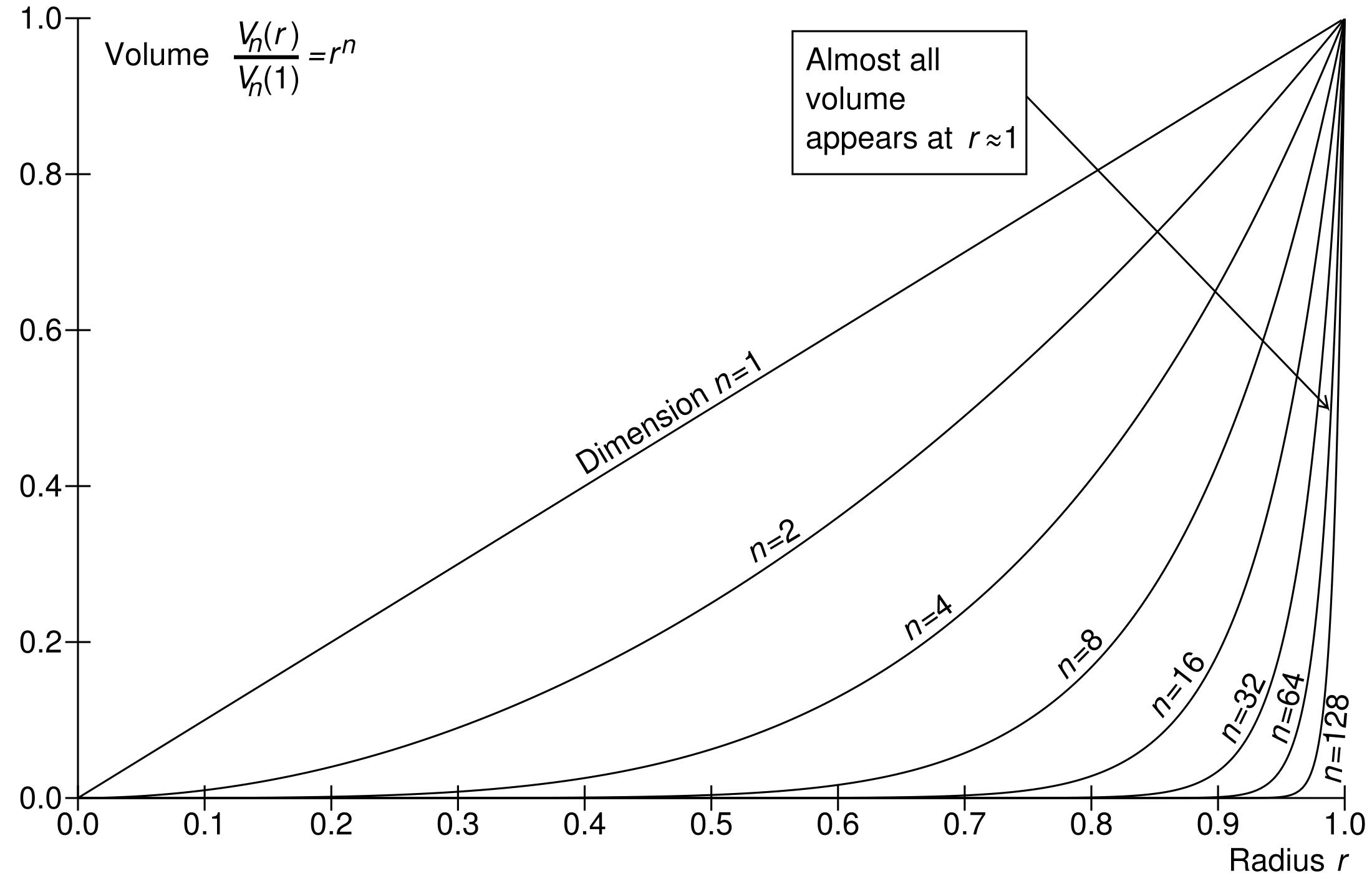}
\caption {\label{Volume}Volume of a high-dimensional ball is concentrated near its border (sphere). Here, $V_n(r)=r^nV_n(1)$ is the volume of the $n$-dimensional ball of radius $r$.}
\end{centering}
\end{figure}

Measure concentration phenomena were discovered in statistical physics. The Gibbs theorem about equivalence of ensembles is one of the first results in this direction: the microcanonical ensemble (equidistribution on an isoenergetic surface in an invariant measure, that is, the phase volume in an infinitesimally thin layer) is equivalent to the canonical distribution that maximizes the entropy \cite{Gibbs1902} under some assumptions about regularity of the interaction potential. The simplest geometric particular case of this theorem is: equidistribution in a ball is equivalent both to the equidistribution on the sphere (Fig.~\ref{Volume}) and to the Gaussian distribution with the same expectation of $r^2=\sum_i x_i^2$, where $x_i$ are the coordinates.

Maxwell used the {concentration of measure} phenomenon to prove the famous Maxwell distribution. Consider a rotationally symmetric probability
distribution on the $n$-dimensional unit sphere. Then, its orthogonal projection on a line is close to the Gaussian distribution with small variance $1/n$ (for large $n$, with high accuracy).
This is exactly the Maxwellian distribution for one degree of freedom in a gas. The distribution on the unit sphere is here the microcanonical distribution of the kinetic energy of a gas, when the potential energy is negligible.   Geometrically, this means that the `observed diameter' in one-dimensional projections of the unit sphere is small. It is of the order of $1/\sqrt{n}$.

L\'evy \cite{Levy1951} took this phenomenon very seriously. He was the first mathematician who recognised  it as a very important property of the geometry of multidimensional spaces. He noticed that instead of orthogonal projections on a straight line  we can use any $\eta$-Lipschitz function $f$ (with $\|f(x)-f(y)\|\leq \eta\|x-y\|$).
Let points $x$ be distributed on a unit $n$-dimensional sphere with rotationally symmetric probability distribution. Then, the values
of $f$ will be distributed `not more widely' than a normal distribution around the  {\em median value} of $f$, $M_f$: for all $\varepsilon>0$
$$\mathbf{P}(|f-M_f|\geq \varepsilon)\leq 2\exp \left(-\frac{n\varepsilon^2}{2c\eta^2}\right),$$
where $c$ is a constant, $0<c\leq 1$. From the statistical mechanics point of view, this L\'evy Lemma
describes the upper limit of fluctuations in a gas for an arbitrary observable quantity $f$. The only condition is {the} sufficient regularity of $f$ (Lipschitz property).

Khinchin created mathematical foundations of statistical mechanics based on the central limit theorem and concentration inequalities \cite{Khinchin1949}. This is one of several great `Mathematical Foundations' aimed to answer the challenge of Hilbert's 6th problem  \cite{GorTyukPhil2018}.

`Blessing of dimensionality' is the recently coined term for situations where high-dimensional complexity makes computation easier \cite{Kainen1997}. It is opposite to the famous `curse of dimensionality'. The curse and  blessing of dimensionality are two sides of the same coin.

Two unit random vectors in high dimensional space are almost orthogonal with high probability. This is a simple manifestation of the so-called { waist concentration} \cite{Gromov2003}. Nearly all area of a high-dimensional sphere is concentrated near its equator. This is obvious: just project a sphere onto a hyperplane and use the concentration argument for a ball on the hyperplane (with a simple trigonometric factor). This seems to be highly non-trivial if we ask: near which equator? The answer is obvious but counter-intuitive: near each equator. There are exponentially large (in dimension $n$) sets of almost orthogonal vectors (with small value of inner products) in $n$-dimensional Euclidean space \cite{Kurkova1993}. This `quasiorthogonality' was used for indexing in high-dimensional data bases \cite{Hecht-Nielsen1994}. Applications of quasiorthogonality to data separation problems are presented in Section `Quasiorthogonal sets and Fisher separability of not i.i.d. data' of \cite{GorbanGolubGrechTyu2018}.

Moreover, exponentially large numbers of randomly and independently chosen vectors from equidistribution on a sphere (and from many other distributions) are almost orthogonal with probability close to one \cite{GorbTyuProSof2016}. The probabilistic approach to quasiorthogonality is proven to be useful  for construction of dimensionality reduction by random mapping \cite{Kaski1998,RitterKohonen1989}.

The classical {concentration of measure} theorems state that independent identically distributed (i.i.d.) random points are concentrated in a thin layer near a surface. This layer can be a sphere or an equator of a sphere, an average or median level set of energy or another Lipschitz function, etc.

The novel { stochastic separation theorems} describe the fine structure of these thin layers \cite{GorbTyu2017}: each point $\boldsymbol{x}$ from a finite  random set $\boldsymbol{X}$ is linearly separable from $\boldsymbol{X}\setminus \{\boldsymbol{x}\}$ with high probability, even for exponentially large random sets $\boldsymbol{X}$. The linear functionals for separation of points can be selected in the form of the simplest linear Fisher's discriminant. For example, $2.7\times 10^6$ points $\boldsymbol{x}_1,\ldots , \boldsymbol{x}_{2,700,000}$ sampled independently from the equidistribution in the 100-dimensional unit ball have the following property with probability $p>0.99$: all $\|\boldsymbol{x_i}\|>1/\sqrt{2} $ and
$$(\boldsymbol{x_i},\boldsymbol{x_j})<\|\boldsymbol{x_i}\|/\sqrt{2} \mbox{ for all } i,j=1,\ldots, 2.7\times 10^6, \, i\neq j,$$
where $(\cdot,\cdot)$ is the standard inner product and $\|\cdot\|$ is the Euclidean norm.

Stochastic separation theorems hold for a much wider class of probability distributions  than just equidistributions in a   ball or Gaussian distributions.  The requirement of i.i.d. samples can  be significantly relaxed \cite{GorbanGrechukTykin2018}. Other simple decision rules can be used instead of Fisher's linear discriminants. For example, simple radial basis functions and Gaussian mixtures can also be easily implemented for separation of points in high dimensions  \cite{AndersonEtAl2014}. In  most  practical situations we have encountered, linear separation seems to be more useful than radial basis functions, since linear discriminants combine the ability of stochastic separation and  the reasonable ability of generalization.

 The main requirement for a data point distribution is that it must be significantly multidimensional. Some effort was needed to answer the question: what distributions are essentially multidimensional in order to guarantee the stochastic separation properties? A special case  of this question was formulated in 2009 by Donoho and Tanner as an open problem \cite{DonohoTanner2009}.
The answer \cite{GorbanGrechukTykin2018} is based on the idea: the sets of  relatively  small volume should not have  relatively  high probability, with detailed specification of these `relatively small' and `relatively high' in the conditions of the theorems.

Thus, `high-dimensional data' does not simply mean `data with many coordinates'. For example, data, located  on a straight line, low-dimensional plane, or a smooth curve with bounded (and not very large) curvature  in high dimensional space are not high-dimensional data. Then,  a special preprocessing is needed to prepare the data for application of simple separating elements, like threshold elements or single neurons. In this preprocessing, the data should be compressed to their {\em genuine dimension}. Stochastic separation is possible if the genuine dimension is sufficiently high. This preprocessing and the evaluation of the genuine dimension requires special techniques. In machine learning it is a combination of various linear and nonlinear dimensionality reduction methods  \cite{GorbanKegl2008,GorZin2010},  compressed sensing \cite{Donoho2006,Eldar2012}, autoencoders \cite{Kramer1991,VincentBengio2008} and other approaches. To the best of our knowledge, this preprocessing cannot be performed by ensembles of simple noninteracting elements.

Thus, in high dimensional data flow, {\em many concepts can be separated by simple elements} (or neurons).
This result gives grounds for the convergent evolution of   artificial and natural intelligence despite  very different timescale of their changes. They use simple elements in a similar way.

One can expect that a  system of perception and recognition consists of two subsystems. The first subsystem preprocesses data, compresses them and extracts features. The second subsystem recognises patterns and returns the relevant concepts. The preprocessing system is expected to be strongly connected, while the recognition system can consist of simple noninteracting elements (grandmother cells). These elements receive preprocessed data and give the relevant response. The blessing of dimensionality phenomena are utilised by the second system, whose elements are simple, do not interact, and recall the concept cells \cite{AndersonEtAl2014}.

A more general architecture was proposed recently for AI systems. The layer  of noninteracting  `concept cells' or the cascade of such layers grows as  a system for correction of errors of a legacy intellectual system.
Any AI system makes errors. The successful functioning of a  system in realistic operational conditions dictates that the errors should be detected and corrected immediately. Real-time correction of  mistakes by re-training is not always viable due to the resources involved. Moreover, the re-training could introduce new mistakes and damage existing skills. Correction of   errors is  becoming an increasingly important and challenging problem in AI technology.

\begin{figure}
\centering
\includegraphics[width=0.7\textwidth]{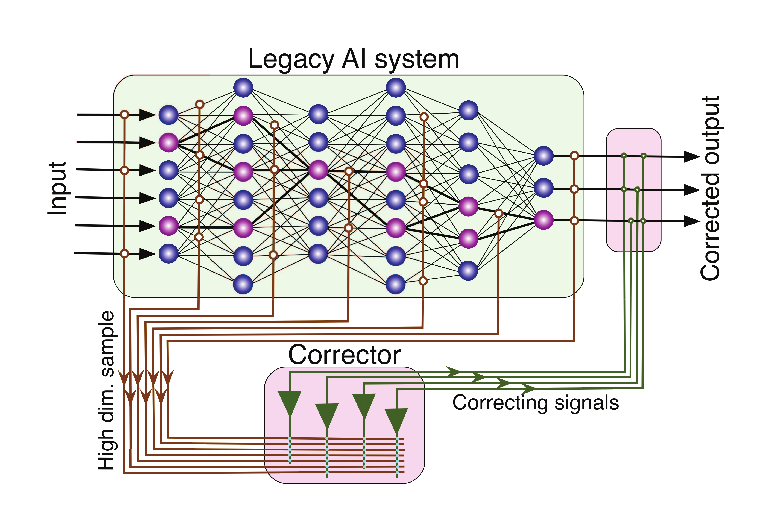}
\caption {Corrector of AI errors. The legacy AI system is represented as a network of elementary units. Inputs for this corrector may include input signals, and any internal or output signal of the AI system (marked by circles).}
\label{Fig:Corrector}
\end{figure}

The recently proposed architecture of such an elementary corrector   \cite{GorbanTyuRom2016} consists of two ideal devices (Fig.~\ref{Fig:Corrector}):
\begin{itemize}
\item A binary classifier for separation of the situations with possible mistakes from the situations with correct functioning (or, more advanced, separation of  the situations with high risk of mistake from the situations with low risk of mistake);
\item A new decision rule for the situations with possible mistakes (or with high risk of mistakes).
\end{itemize}

\begin{figure} 
\centering
\includegraphics[width=0.75\textwidth]{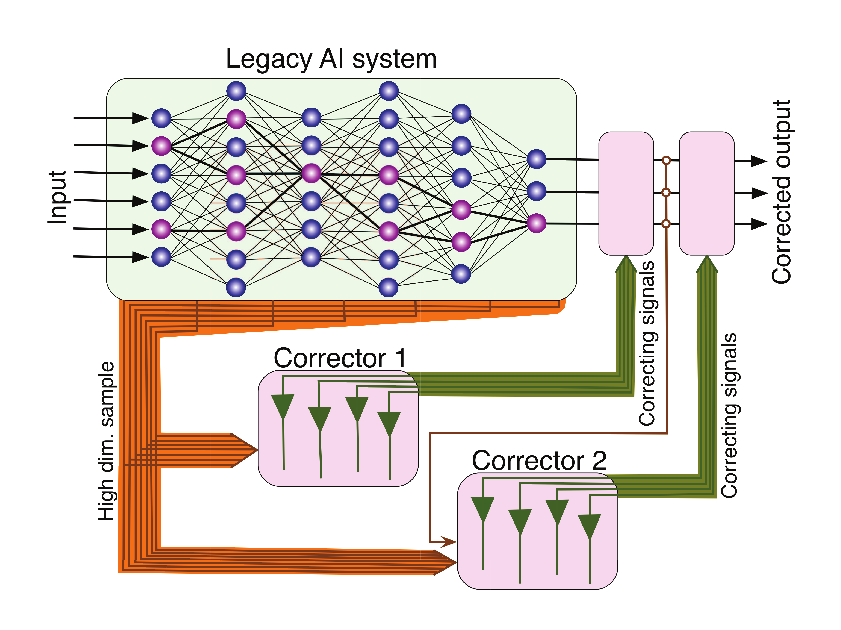}
\caption{Cascade of AI correctors. In this diagram, the original legacy AI system (shown as Legacy AI System 1) is supplied with a corrector altering its responses.  The combined new AI system can in turn be augmented by another corrector, leading to a cascade of AI correctors. }\label{Fig:AIcorrectorsCascade}
\end{figure}

One corrector can fix several errors (it is useful to cluster them before corrections). Several correctors can work independently with  low probability of conflict. Cascades of correctors are employed for further correction of more errors \cite{GorTyukPhil2018}: the AI system with the first corrector is a new legacy AI system and can be corrected further (Fig.~\ref{Fig:AIcorrectorsCascade}).

Fascinating speculations and hypotheses about correctors in the evolution of the brain can be
suggested, but here we avoid this topic and try not to deviate significantly from strictly proven theorems or results supported by experiments or computational experiments. Effectiveness of one-trial correctors of AI errors is supported both by the theorems and the computational experiments. Existence and selectivity of concept cells has been proven experimentally. Thus,  we observe the {\em simplicity revolution} in neuroscience that started from invention of grandmother cells and approached recently the theory and technology of one-short learning of AI correctors.

In Sec.~\ref{Sec:GrandMother} we review the results about grandmother cells, concepts cells and space coding. Blessing of dimensionality and stochastic separation theorems are presented in Sec.~\ref{Sec:StochSep}. Computational experiments with correctors of legacy AI systems, knowledge transfer between  AI systems using correctors, and cascades of correctors are discussed in Sec.~\ref{Sec:Correctors}. In Sec.~\ref{Sec:BMBLearning} we return to brain dynamics and present a model of appearance of codifying memory in stratified brain structures such as  the hippocampus following \cite{TyukinBrain2017}.  We show how single neurons can selectively detect and learn arbitrary information items, given that they operate in high dimensions.  These results constitute a basis for organization of complex memories in ensembles of single neurons.

\section{Grandmother cells,  concept cells, and specific coding \label{Sec:GrandMother}}

Three years after the term `grandmother cell' was proposed,  Barlow published an essay about the single neuron revolution  in sensory psychology \cite{Barlow1972}. His central proposition was:  `our perceptions are caused by the activity of a rather small number of neurons selected from a very large
population of predominantly silent cells.' It was described in more detail in the `five dogmas' of a single cell perception:
\begin{enumerate}
\item To understand nervous function one needs to look at interactions at a cellular level, rather than
either a more macroscopic or microscopic level, because behaviour depends upon the organized
pattern of these intercellular interactions.
\item The sensory system is organized to achieve as complete a representation of the sensory stimulus
as possible with the minimum number of active neurons.
\item Trigger features of sensory neurons are matched to redundant patterns of stimulation by
experience as well as by developmental processes.
\item Perception corresponds to the activity of a small selection from the very numerous high-level
neurons, each of which corresponds to a pattern of external events of the order of complexity of
the events symbolized by a word.
\item High impulse frequency in such neurons corresponds to high certainty that the trigger feature is
present.
\end{enumerate}

He presented a collection of examples of single cell perception found experimentally.	In all these examples  neurons reacted selectively to the key patterns (called  'trigger
features') and this reaction was invariant to the change of conditions like variations in light intensity or presence of noise. Such sensory neurons can be located in various places, for example, in retina or in mammalian cerebral cortex. The trigger features of the retinal ganglion cells are the key patterns of the visual information. They are more elementary than the objects or concepts  (like `grandmother' or `jedi'). The  higher order concepts are combined from these trigger features. Therefore, the representation of objects and concepts by the retina can be considered as distributed or implicit. They are coded explicitly at higher level of the visual information processing stream \cite{QuianQuirogaKreiman2010}.

The status of these five dogmas was defined as a simple set of hypotheses  that were compatible with known facts. Barlow also criticised the concept of grandmother cell because it neglects the connection between one perception and others. He suggested that unique events are not isolated in our perception, they overlap with each other and the concept of  grandmother cell makes this continuity impossible.

Discussion about the role of single cells in perception has a long history. James in his book \cite{James1890} (Chapter VI, p. 179) considered the possibility to revive Leibniz's multiple monadism (or polyzoism):  `Every brain-cell has its own individual consciousness, which no other cell knows anything about, all individual consciousness being 'ejective' to each other. There is, however, among the cells one central or pontifical one to which {\em our} consciousness is attached' (following the term `grandmother cell', we can call the pontifical cell the `I-cell'). Sherrington criticises the idea of pontifical cell and proposed the `million-fold democracy whose each unit is a cell' \cite{Sherrington1941}. He stated that the perception of any specific object is performed by the joint activity of many millions of neurons.

Barlow proposed the concept of `cardinal cells'. This is the idea of sparse coding: `Among the many cardinals only a few speak at once'  \cite{Barlow1972}. He also attracted attention of readers to the fact that the number of cortical neurons in area 17
is orders of magnitude greater than the number of incoming fibres (area 17, or V1, or striate cortex  is the end organ of the afferent visual system and is situated in the occipital lobe). This can be considered as a sign that there are many more neurons for codifying memory and perception than the dimension of the visual inputs into this area. The hierarchy of sensory neurons is very different from the church hierarchy: there are many more  cardinal cells than the ordinary `church members'.

A series of experiments demonstrated  that neurons in the human medial temporal lobe (MTL) fire selectively to images of faces, animals, and other objects or scenes \cite{Fried1997,Kreiman2000,QuianQuirogaNature2005}. It was demonstrated that the firing of MTL cells was sparse because most of them did not respond to the great majority of images used in the experiment. These cells have low baseline activity and their response is highly selective \cite{QuianQuirogaNature2005}. `Jennifer Aniston' cells responded to pictures of Jennifer Aniston but rarely of very weekly to pictures of other persons. An important observation was: these neurons respond also to the printed name of the person. Therefore, they react not only on the visual likeness but on the concept. Moreover, the voluntary control of these neurons is possible via imagination: it is sufficient  to imagine the concept or `continuously think of the concept' \cite{Cerf2010}.

These discoveries clearly demonstrate fantastic selectivity of the neuronal response. At the same time, the `ideal' grandmother cells with one-to-one relations between the concept cells and the concepts were not found:
\begin{itemize}
\item There are several concept cells for each concept. A single cell is impossible to find, therefore, the fact that these cells were found proved that there were sufficiently many `Jennifer Aniston' cells, `Brad Pitt' cells, etc.
\item One cell can fire for different concepts -- this is association between concepts. For example, the `Jennifer Aniston' cells also fired to Lisa Kudrow, a costar in the TV series Friends, and `Luke Skywalker' cell also fired to Yoda \cite{QuianQuirogaSciAm2013}. This means that the concept extracted by these units are wider than just single name or character. Nevertheless, the presence of the individual can, in most cases, be reliably decoded from a small number of neurons.
\item Redistribution of attention between different parts of the image can have effect opposite to association. For example, the `Jennifer Aniston' cell did not react on the picture, where Jennifer Aniston is together with Brad Pitt. It is recognised rather as Brad Pitt.
\end{itemize}
One can modify the concepts behind the images and insist that there   {are} proper `grandmother cells' (may be several for each concept) and we should just properly define the concept of `grandmother'. For example, the units, which react both on Luke Skywalker and Yoda, can be called the `Famous Jedi' cells, whereas the units, which fire both to Jennifer Aniston and Lisa Kudrow are `The stars of the TV series Friends' cells. There was an intensive discussion about the definition of the  proper grandmother cells \cite{Bowers2009,Bowers2010,Plaut2010,QuianQuirogaKreiman2010,Roy2013}, which was ended by the analysis of the notion, that was necessary because `a typical problem in any discussion about grandmother cells is that there is not a general consensus about what should be called as such' \cite{QuianQuirogaKreiman2010a}.

A `neuron-centred' idea of concepts is possible: a concept is   {a} pattern of the input information that   {evokes} a specific selective response of a group of cells. In this sense, the grandmother cells exist. Conventional notion of `concepts' is different: these `concepts' are used in communication between people and   {as such} are the compromises between   {neuron-centred concepts in} different individuals.   {Convergence} of these different worlds of concepts is also possible because the concept cells   {themselves} can learn rapidly.   {Timescales} of this learning is compatible with   {that of the} episodic memory formation.   {Classical experiments revealing} such learning in hippocampal neurons   {were} presented  in 2015 \cite{IsonQQ2015}.
A cell was found that  actively responded to presentation of the image of the patient's family and did not respond to the image of the Eiffel tower before learning. In the single trial learning the composite picture, the family member at the Eiffel tower, was exposed once. After that, the firing rate of the cell in response to the Eiffel
tower increased significantly,  the response to the image of the family member did not change significantly, and it remained similar to the response to the composite image. This plasticity and rapid learning of associations do not fully correspond to the `monadic' idea of the grandmother's cell or to  the idea of `concept cell' with rigorously defined concepts. The discussion about terminology and  interpretation could continue further.

Nevertheless, without any doubt and without re-definition of the concepts,  the experiments with single-cell perception show that there are (relatively) small sets of neurons that react actively and selectively on the input images associated with a small group of interrelated concepts (and do not react actively on other images). That is, the small neural ensembles separate the images, associated with a group of associated concepts, from other images.

There are several other paradigms of coding of sensory information in brain. Distributed coding as   opposite  to the local one is used   for coding and prediction of motion \cite{Eichenbaum1993,Wilson1993}. Correlation between different neurons are important for information decoding at the cell population level even if the pair correlations are rather low  \cite{Averbeck2006}.  The balance between single cell  and distributed coding  remains a non-trivial problem in neuroscience \cite{Valdez2015}. The intermediate paradigm of sparse coding suggests   {that} sensory information is encoded using a small number of neurons active at any given time  \cite{Olshausen2004} (only a few cardinals speak at once). Sparse coding decreases `collisions'
(less intersections) between patterns. It is also beneficial  for learning associations. According to Barlow \cite{Barlow1972}, the nervous system  tends to form neural coding with higher degrees of specificity, and    {sparse coding is just an intermediate step} on the way to   {achieve} maximal specificity.

In the next section we demonstrate, why highly specific coding by simple elements (concept cells) is possible and beneficial for high-dimensional brain in high-dimensional world and explain what  `high-dimensional' means in this context.

\section{Blessing of dimensionality: concentration of measure and stochastic separation theorems \label{Sec:StochSep}}

In the sequel, we use the following notations:  $\Real^n$ is the $n$-dimensional linear real vector space, $\boldsymbol{x} =(x_{1},\dots,x_{n})$ denote elements of $\Real^n$,  $(\boldsymbol{x},\boldsymbol{y})=\sum_{k} x_{k} y_{k}$ is the standard inner product of $\boldsymbol{x}$ and $\boldsymbol{y}$, and $\|\boldsymbol{x}\|=\sqrt{(\boldsymbol{x},\boldsymbol{x})}$ is the  Euclidean norm  in $\Real^n$. Symbol $\mathbb{B}_n$ stands for the unit ball in $\Real^n$ centred at the origin, $\mathbb{B}_n=\{\boldsymbol{x}\in\Real^n| \ \left(\boldsymbol{x},\boldsymbol{x}\right)\leq 1\}$.  $V_n$ is the $n$-dimensional Lebesgue measure, and $V_n(\mathbb{B}_n)$ is the volume of a unit ball.  $\mathbb{S}^{n-1}\subset \mathbb{R}^{n}$ is the unit sphere in $\mathbb{R}^{n}$. For a finite set $Y$, the number of points in $Y$ is $|Y|$.

\subsection{Fisher's separability of a point from a finite set}

Separation of a point from a finite set is a basic  task in data analysis. For example, for correction of a mistake of an AI system we have to separate the situation with mistake from a relatively large set of situations, where the system works properly.

In high dimensions simple linear Fisher's discriminant demonstrates the surprisingly high ability to separate points. After data {\em whitening}, Fisher's discriminant can be defined by a simple linear inequality:
\begin{definition}
A point $\boldsymbol{ x}$ is Fisher separable from a finite set $Y $ with a threshold $\alpha$ ($0\leq  \alpha<1$) if 
\begin{equation}\label{discriminant}
(\boldsymbol{x},\boldsymbol{y})\leq \alpha (\boldsymbol{x},\boldsymbol{x}),
\end{equation}
 for all $\boldsymbol{y}\in Y$.
 \end{definition}
Whitening is a special change of coordinates that transforms the empiric covariance matrix into identity matrix. It can be represented in four steps:
 \begin{enumerate}
 \item Centralise the data cloud (subtract the mean from all data vectors), normalise the coordinates to unit variance and calculate the empiric correlation matrix.
  \item Apply principal component analysis (i.e. calculate eigenvalues and eigenvectors of empiric correlation matrix). 
 \item Delete  minor components, which correspond to the small eigenvalues of empiric correlation matrix.
\item  In the remained principal component basis normalise coordinates to unit variance.
\end{enumerate}

Deletion of minor components is needed to avoid multicollinearity, i.e. strong linear dependence between coordinates.  Multicollinearity makes the models sensitive to fluctuations in data and unstable. The condition number of the correlation matrix is the standard measure of multicollinearity, that is the ratio $\kappa=\lambda_{\max}/\lambda_{\min}$, where $\lambda_{\max}$ and $\lambda_{\min}$ are the maximal and the minimal eigenvalues of this matrix. Collinearity is strong if  $\kappa>10$. If  $\kappa<10$ then collinearity is considered as modest and most of the  standard regression and classification methods work reliably \cite{Dormann2013}.

Whitening has linear complexity in the number of datapoints. Moreover, for approximate whitening we do not need to use all data and sampling makes complexity of whitening essentially sublinear.

It could be difficult to provide the precise whitening for intensive real life data streams. Nevertheless, a rough approximation to this transformation creates useful discriminants (\ref{discriminant}) as well.  With some extension of  meaning, we call `Fisher's discriminants'  all the linear discriminants created non-iteratively by inner products (\ref{discriminant}).

\begin{definition}A finite set $F \subset {\mathbb R}^n$ is called \emph{Fisher-separable} with threshold $\alpha \in (0,1)$ if inequality (\ref{discriminant})
holds for all $\boldsymbol{ x}, \boldsymbol{ y} \in F$ such that $\boldsymbol{ x}\neq  \boldsymbol{ y}$. The set $F$ is called \emph{Fisher-separable} if there exists some $\alpha $ ($0\leq  \alpha<1$)  such that $F$ is Fisher-separable with threshold $\alpha$.  
\end{definition}

\begin{figure}
\centering
\includegraphics[width=0.3\textwidth]{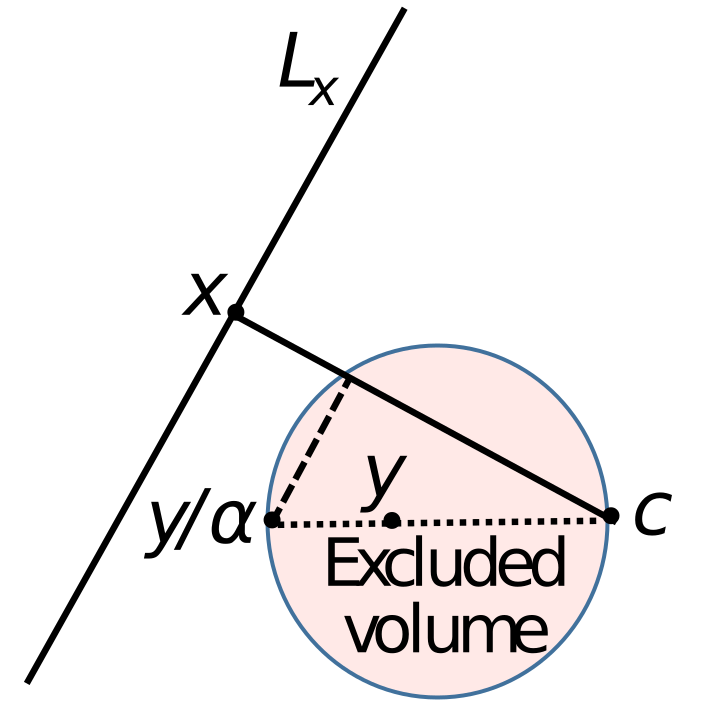}
\caption{Diameter of the filled ball (excluded volume) is the segment $[\boldsymbol{c},\boldsymbol{y}/\alpha]$. Point $\boldsymbol{x}$ should not belong to the excluded volume to be separable from $\boldsymbol{y}$ by the linear discriminant (\ref{discriminant}) with threshold $\alpha$. Here, $\boldsymbol{c}$ is the origin (the centre), and $L_x$ is the hyperplane such that $(\boldsymbol{x},\boldsymbol{z})=(\boldsymbol{x},\boldsymbol{x})$ for $\boldsymbol{z} \in L_x$.  A point $\boldsymbol{x}$ should not belong to the union of such balls for all $\boldsymbol{y} \in Y$ for separability from a set $Y$.}
\label{Fig:Excluded}
\end{figure}

Using elementary geometry, we can show  that  inequality (\ref{discriminant}) holds for vectors $\boldsymbol{x}$, $\boldsymbol{y}$ if and only if $\boldsymbol{x}$ does not belong to a ball (Fig.~\ref{Fig:Excluded}):
\begin{equation}\label{excludedvolume}
\left\{\boldsymbol{z} \ \left| \ \left\|\boldsymbol{z}-\frac{\boldsymbol{y}}{2\alpha }\right\|< \frac{\|\boldsymbol{y}\|}{2\alpha} \right.  \right\}.
\end{equation}
The volume of  ball (\ref{excludedvolume}), where the discriminant inequality (\ref{discriminant}) does not hold,   can be relatively small. 
For example, if $Y$ is a subset of $\mathbb{B}_n$, then the volume of each ball (\ref{excludedvolume}) does not exceed $\frac{1}{(2\alpha)^n} V_n(\mathbb{B}_n)$. To use further this simple estimate assume that $\alpha>1/2$. Point $\boldsymbol{x}$ is separable from a set $Y$ by Fisher's linear discriminant with threshold $\alpha$ if it does not belong to the union of these excluded balls. The volume of this union does not exceed
\begin{equation}\label{ExclVol}
V_{\rm excl}\leq \frac{|Y|}{(2\alpha)^n} V_n(\mathbb{B}_n).
\end{equation}
If $|Y|<b^n$ with $1<b < 2 \alpha$ then the fraction of excluded volume in the unit ball decreases exponentially with dimension $n$ as $\left(\frac{b}{2\alpha}\right)^n$.

 \begin{proposition}\label{Prop:ExclVol}Let $0<\psi<1$, $Y\subset \mathbb{B}_n$ be a finite set, $|Y|<\psi (2\alpha)^n$, and $\boldsymbol{x}$ be a randomly chosen point from the equidistribution in the unit ball. Then with probability $p>1-\psi$ point $\boldsymbol{x}$ is Fisher-separable from $Y$ with threshold $\alpha$ (\ref{discriminant}).
\end{proposition}
This proposition is valid for any finite set  $Y\subset \mathbb{B}_n$ and without any hypotheses about its statistical nature. No underlying probability distribution is assumed for points of $Y$. For $\boldsymbol{x}$, Proposition  \ref{Prop:ExclVol} assumes equidistribution in a ball. Of course, the class of distributions satisfying this property is much larger. It is required that the probability to find $\boldsymbol{x}$ in a small volume $V_{\rm excl}$ (\ref{ExclVol}) is small and vanishes when dimension $n \to \infty$. Assume that the number of points $|Y|$ can grow with dimension and is bounded by a geometric progression, $|Y|<b^n$, $b<2\alpha$.  The simplest condition on the distribution of   $\boldsymbol{x}$ can be formulated for the probability density. Let the probability distribution in a unit ball  $\mathbb{B}_n$ have the continuous density $\rho$, and $\rho_m=\max\{\rho(\boldsymbol{x})\}$. Then the probability to find a random point   $\boldsymbol{x}$ in the excluded volume  $V_{\rm excl}$ (\ref{ExclVol})  does not exceed
$$\psi=\rho_m V_{\rm excl}\leq \rho_m  \left(\frac{b}{2\alpha}\right)^n V_n(\mathbb{B}_n).$$
For the equidistribution in a ball, $\rho_m=\rho=1/V_n(\mathbb{B}_n)$. Consider the distributions that can deviate significantly from the equidistribution, and these deviations can grow with dimension but   not faster than the geometric progression with the common ratio $1/r>1$:
\begin{equation}\label{bounded}
\rho_m<\frac{C}{r^n V_n(\mathbb{B}_n)},
 \end{equation}
here  $C$ does not depend on $n$.

For such a distribution in the unit ball, the probability $\psi$ to find a random point   $\boldsymbol{x}$ in the excluded volume  $V_{\rm excl}$ (\ref{ExclVol}) tends to 0 when $n\to \infty$ as a geometric progression with the common ratio ${b}/({2r\alpha})$.  We now formulate this important fact as a theorem.
\begin{theorem}\label{Theorem:ExclVol2}
Assume that a probability distribution in the unit ball has a density with maximal value $\rho_m$, which satisfies inequality (\ref{bounded}). Let $|Y|<b^n$ and $2r\alpha>b>1$.
Then  the probability $p$ that a random point is Fisher-separable from the finite set $Y$ is $p=1-\psi$, where $$\psi<C\left(\frac{b}{2r\alpha}\right)^n.$$
\end{theorem}

Two restriction  on the distribution of $\boldsymbol{x}$ were used: the absence of large deviations (the distribution has a bounded support contained in a  unit ball) and the boundedness of the density (\ref{bounded}). The first restriction allowed us to consider the balls of excluded volume (Fig.~\ref{Fig:Excluded}) with bounded diameter $d<1/\alpha<2$. The second restriction guarantees that the probability to find a point in such a ball is small. The essence of this restriction is: the sets of relatively small volume have relatively small probability. For the whitened data this means that the distribution is essentially $n$-dimensional and is not concentrated near a low-dimensional nonlinear `principal object', linear or nonlinear one \cite{GorbanKegl2008,GorZin2010,ZinMir2013}.

For analysis of Fisher-separability   for  general distribution, an auxiliary random variable is useful. This is the probability that  a randomly chosen point $\boldsymbol{x}$ is {\em not} Fisher-separable with threshold $\alpha$ from a given data point  $\boldsymbol{y}$ by the discriminant (\ref{discriminant}) (i.e.  the inequality (\ref{discriminant}) is false):
\begin{equation}\label{excluded}
p=p_y(\alpha)=\int_{\left\|\boldsymbol{z}-\frac{\boldsymbol{y}}{2\alpha }\right\|\leq \frac{\|\boldsymbol{y}\|}{2\alpha}} \rho(\boldsymbol{z}) \,d\boldsymbol{z} ,
\end{equation}
where $ \rho(\boldsymbol{z}) \,d\boldsymbol{z}$ is the probability measure for  $\boldsymbol{x}$.
For example,   for the equidistribution in a ball $\mathbb{B}_n$ and arbitrary  $\boldsymbol{y}$ from the ball, $\rho(\boldsymbol{z})=1/V_n(\mathbb{B}_n)$ and $p_y <\left(\frac{1}{2\alpha}\right)^n$. The probability to select a point  {inside} the union of $N$ `excluded balls'  is less than $N\left(\frac{1}{2\alpha}\right)^n$ for any sampling of $N$ points $\boldsymbol{y}$ in $\mathbb{B}_n$.

 If $\boldsymbol{y}$ is a random point (not compulsory with the same distribution as $\boldsymbol{x}$) then $p_y(\alpha)$ is also a random variable. For a finite set of data points $Y$ the probability $p_{Y}(\alpha)$ that the data point is {\em not}  Fisher-separable with threshold $\alpha$ from the set $Y$ can be evaluated by the sum of $p_y(\alpha)$ for $\boldsymbol{y}\in Y$:
\begin{equation}\label{sump_y}
p_Y(\alpha) \leq \sum_{y\in Y} p_y(\alpha).
\end{equation}
Evaluation of sums of random variables (dependent or independent, identically distributed or not) is a classical area of probability theory \cite{Zolotarev2011}. The inequality (\ref{sump_y}) in combination with the known results provide many versions and generalizations  of Theorem~\ref{Theorem:ExclVol2}. For this purpose, it is necessary to evaluate the distribution of $p_y(\alpha)$ and its moments. 

Evaluation of  the distribution of $p_y(\alpha)$ and its moments does not require the knowledge  the detailed  multidimensional distribution of  $\boldsymbol{y}$. It can be performed over an empirical dataset as well. Comparison of the empirical distribution of  $p_y(\alpha)$ with the distribution evaluated for the high-dimensional sphere $\mathbb{S}^{n-1}\subset \mathbb{R}^{n}$  provides an interesting information about the `genuine' dimension of data. The probability $p_y(\alpha)$ is the same for all $y \in \mathbb{S}^{n-1}$ and exponentially decreases for large $n$.

Simple estimate gives for the rotationally invariant  equidistribution on the unit sphere  $\mathbb{S}^{n-1}\subset \mathbb{R}^{n}$  \cite{GorbanGolubGrechTyu2018}:
\begin{equation}\label{p_y on sphere}
p_y(\alpha)\approx   \frac{(1-\alpha^2)^{(n-1)/2}}{\alpha \sqrt{2\pi (n-1)}}.
\end{equation}
Here $f(n)\approx g(n)$ means  that $f(n)/g(n)\to 1$ when $n\to \infty$ (note that the functions here are strictly positive).

\subsection{ A LFW case study}\label{Sec:test}

In this Section, we evaluate the moments of $p_y(\alpha)$ for a well known benchmark database  LFW (Labeled Faces in the Wild). It is a set of images of famous people: politicians, actors, and singers \cite{LFWSurvey2016}. There are 13,233 photos of 5,749 people in LFW, 10,258 images of men and 2,975 images of women. The data are available online \cite{LFW}.

In preprocessing, photos were cropped and aligned. Then, the neural network system  FaceNet was used to calculate 128 features for each image. FaceNet was specially trained to map face images into 128-dimensional feature space \cite{FaceNet2015}. The training was organised to decrease distances between the feature vectors for images of the same person and increase the  distances between the feature vectors  for images of different persons. After this step of preprocessing, the 128-dimensional feature vectors were used for analysis instead of the initial images. In this analysis we follow \cite{GorbanGolubGrechTyu2018}.

\begin{figure}
\centering
\includegraphics[width=0.45\textwidth]{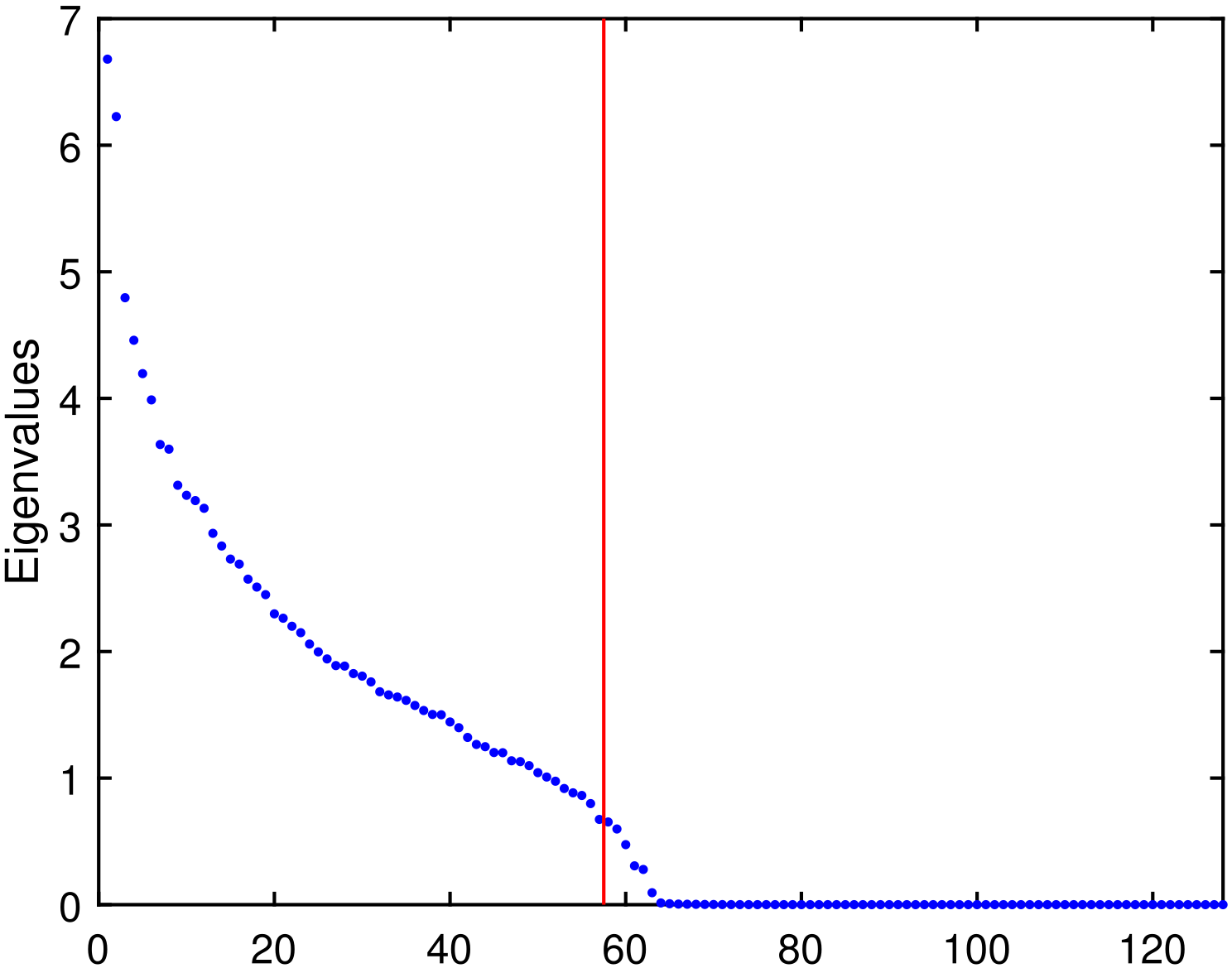}
\caption{Eigenvalue of the correlation matrix for the LFW benchmark as a function of the number of principal component. The verticaL line separates first 57 eigenvalues.}
\label{variance}
\end{figure}

The standard  Principal Component analysis (PCA) provided the essential dimension decrease.  Dependence of the eigenvalue of the correlation matrix $\lambda_k$ on $k$ is presented in Fig.~\ref{variance}.

There exist various heuristics for selection of principal components to retain \cite{Cangelosi2007}. The plot on Fig.~\ref{variance} shows  that not more than 63 principal components  explain practically 100\% of data variance. We applied multicollinearity control for selection of main components.  The recommended border of modest collinearity is $\kappa<10$, where  $\kappa=\lambda_{\max}/\lambda_{\min}$ and $\lambda_{\max}$ and $\lambda_{\min}$ are the maximal and the minimal eigenvalues of the correlation matrix \cite{Dormann2013}. Thus, we retained 57 principal components with the eigenvalues of covariance matrix $\lambda\leq 0.1\lambda_{\max}$. 

Whitening was applied after projection of feature vectors onto these 57 principal components. After that, the covariance matrix became identity matrix. In the basis of principal components, whitening is a coordinate transformation: $x_i \to x_i/\sqrt{\lambda_i}$, where $x_i $ is the projection of the data vector on $i$th principal component, and $\lambda_i$ is the eigenvalue  of the covariance matrix, corresponding to the $i$th principal component.

Optionally, an additional preprocessing operation, the projection on the unit sphere,
 was applied after whitening. This is  the normalization of each data vector to unit length (just take $x/\|x\|$ instead of $x$).

The results of computation are presented in Table~\ref{tab:separability} \cite{GorbanGolubGrechTyu2018}. Here, $\alpha$ is the threshold in Fisher's discriminant (\ref{discriminant}),  $N_{\alpha}$ is the number of points, which cannot be separated from the rest of the data set by Fisher's discriminant (\ref{discriminant}) with given $\alpha$, $\nu_{\alpha}$ is the fraction of these points in the data set, and $\bar{p}_y$ is the mean value of $p_y$.

\begin{table}[!ht]
\centering
\caption{Separability of data points by Fisher's linear discriminant in the preprocessed LFW data set.}
\label{tab:separability}
\begin{tabular}{|l|c|c|c|c|c|}
\cline{1-6}
$\alpha$ &	0.8&0.9&0.95&0.98&0.99  \\ \hline
\multicolumn{6}{|c|}{Separability from all data} \\ \hline
$N_{\alpha}$   &4058&751&123&26&10 \\\hline
$\nu_{\alpha}$  &0.3067&0.0568&0.0093&0.0020&0.0008\\\hline
$\bar{p}_y$  &9.13E-05&7.61E-06&8.91E-07&1.48E-07&5.71E-08 \\\hline
	\multicolumn{6}{|c|}{Separability from points of different classes}\\\hline
$N_{\alpha}^*$   &55&13&6&5&5 \\\hline
$\nu_{\alpha}^*$  &0.0042&0.0010&0.0005&0.0004&0.0004\\\hline
$\bar{p}_y^*$  &3.71E-07&7.42E-08&3.43E-08&2.86E-08&2.86E-08 \\\hline
\multicolumn{6}{|c|}{Separability from all data on unit sphere} \\ \hline
$N_{\alpha}$   &3826&475&64&12&4 \\\hline
$\nu_{\alpha}$  &0.2891&0.0359&0.0048&0.0009&0.0003\\\hline
$\bar{p}_y$  &7.58E-05&4.08E-06&3.66E-07&6.85E-08&2.28E-08 \\\hline
	\multicolumn{6}{|c|}{Separability from points of different classes on unit sphere}\\\hline
$N_{\alpha}^*$   &37&12&8&4&4 \\\hline
$\nu_{\alpha}^*$  &0.0028&0.0009&0.0006&0.0003&0.0003\\\hline
$\bar{p}_y^*$  &2.28E-07&6.85E-08&4.57E-08&2.28E-08&2.28E-08 \\\hline
\end{tabular}
\end{table}

We cannot expect an i.i.d. data sampling for a `good' distribution to be an appropriate model for the LFW data set. Images of the same person are  expected to have more similarity  than images of different persons. It is expected that the set of data in FaceNet coordinates will be even further from  the  i.i.d. sampling of data, because it is prepared to group images of the same person together and `repel' images of different persons. Consider the property of image to be `separable from images of other persons'. Statistics of this `separability from all points of different classes' by Fisher's discriminant is also presented in Table~\ref{tab:separability}. We call  the point $\boldsymbol{x}$ inseparable from points of different classes if at least for one point
$\boldsymbol{y}$ of a different class (image of a different person)
$(\boldsymbol{x},\boldsymbol{y})> \alpha (\boldsymbol{x},\boldsymbol{x})$.
We use stars in Table~\ref{tab:separability} for statistical data related to separability  from points of different classes (i.e., use the notations $N_{\alpha}^*$, $\nu_{\alpha}^*$, and $\bar{p}_y^*$).  When $\alpha$ approaches 1, both $N_{\alpha}$ and $N_{\alpha}^*$ decay, and the separability of a point from all other points becomes better (see Table~\ref{tab:separability}).

It is not surprising that this separability of classes is much more efficient, with less inseparable points. Projection on the unit sphere also improves separability.

It is useful to introduce some baselines for comparison: what values of  $\bar{p}_y$ should be considered as small or large? Two levels are obvious: for the experiments, where all  data points are counted in the excluded volume, we consider the level $p=1/N_{\rm persons}$ as the critical one, here $N_{\rm persons}$ is the number  of different persons in the database. For the experiments, where only images of different persons are counted in the excluded volume, the value $1/N$ seems to be a good candidate to separate the `large' values of  $\bar{p}_y^*$ from the `small' ones. For LFW  data set, $1/N_{\rm persons}=$1.739E-04 and $1/N=$7.557E-05. Both levels have been achieved already for $\alpha=0.8$. The parameter $\nu_{\alpha}^*$ can be considered for experiments with separability from points of different classes as the separability error. This error is extremely small: already for $\alpha=0.8$ it is less than 0.5\% without projection on unit sphere and less than 0.3\% with this projection. The ratio $(N_{\alpha}-N_{\alpha}^*)/N_{\alpha}^*=(\nu_{\alpha}-\nu_{\alpha}^*)/\nu_{\alpha}^*$ can be used to evaluate   the generalisation ability of the Fisher's linear classifier. The nominator is the number of data points (images) inseparable from some points of the same class (the person) but separated from the points of all other classes. For these images we can say that Fisher's discriminant makes some generalizations. The denominator is the number of data points inseparable from some points of other classes (persons). According to the proposed indicator, the generalization ability is impressively high for $\alpha=0.8$: it is 72.8 for preprocessing without projection onto unit sphere and 102.4  for prepocessing with such projection. For $\alpha=0.9$ it is smaller (56.8 and 38.6 correspondingly) and decays fast with further growing of $\alpha$.

\begin{figure}
\centering
\includegraphics[width=0.9\textwidth]{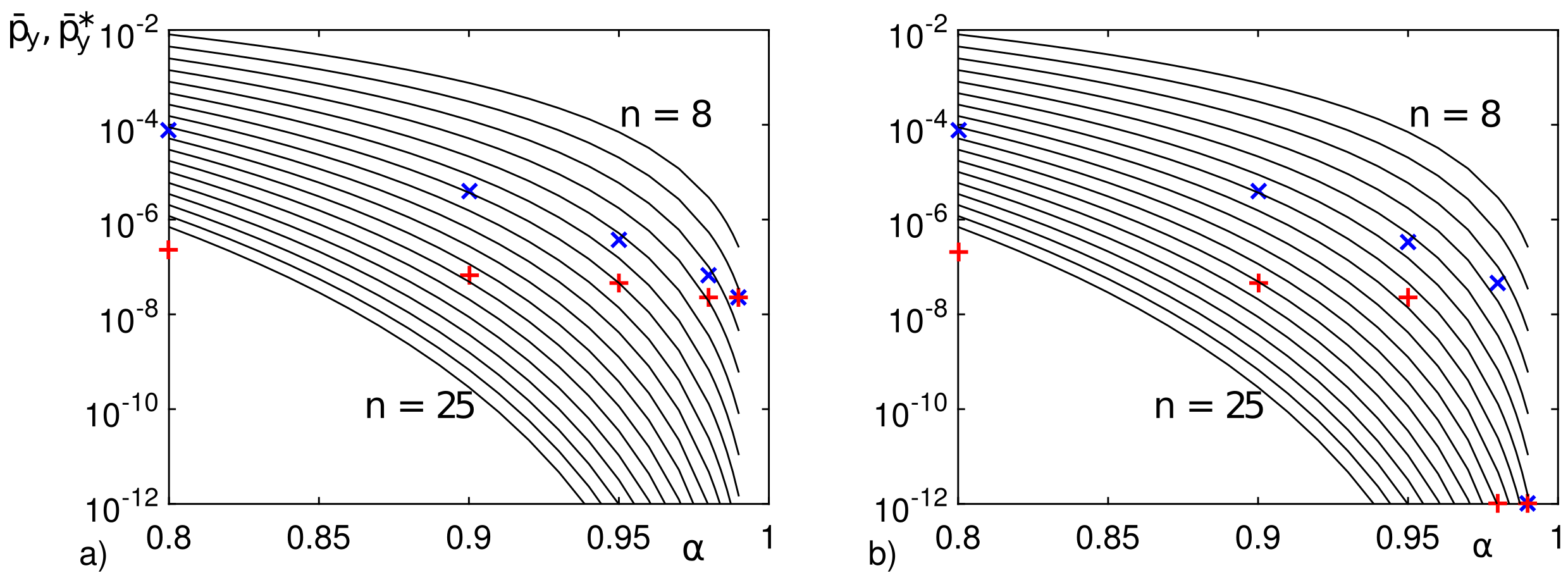} 
\caption{Dependence of $\bar{p}_y$ and  $\bar{p}_y^*$ on $\alpha$. Symbols corresponds to the preprocessed LFW data set projected on the unit sphere (data from Table~\ref{tab:separability}):  pluses, ${\mathbf +}$, are used for  $\bar{p}_y$ and skew crosses, ${\mathbf \times}$, are used for $\bar{p}_y^*$. Solid lines -- dependencies of $\bar{p}_y$ on $\alpha$ for equidistribution on the unite sphere (\ref{p_y on sphere})  in different dimensions, from $n=8$ to $n=25$, from the top down. (a) Original database with saturation at $\alpha=0.99$; (b) The database after fixing two labelling mistakes.  Pluses and crosses on the bottom borderline correspond to zero values.}
\label{Fig:pwithoutlier}

\end{figure}
 
In the projection on the unit sphere, $N_{\alpha},N_{\alpha}^*,\bar{p}_y(\alpha), \bar{p}_y^*(\alpha)\to  0$ when $\alpha \to 1$. This is a trivial separability on the border of a strongly convex body where each point is an extreme one. The rate of this convergence to 0 and the separability for $\alpha<1$ depend on the dimension and on the intrinsic structures in data (multicluster structure, etc., Fig.~\ref{Fig:pwithoutlier}). Compare the observed behaviour of $\bar{p}_y(\alpha)$ and $\bar{p}_y^*(\alpha)$ for the LFW data set projected on the unit sphere to $p_y$ for equidistribution on the unit spere (\ref{p_y on sphere}). We can see in Fig.~\ref{Fig:pwithoutlier}a that the values of $\bar{p}_y(\alpha)$ for the preprocessed LFW database projected on the unit sphere correspond approximately to the equidistribution on the sphere in dimension 16, and for $\alpha=0.9$ this effective dimension decreases to 14. For $\bar{p}_y^*(\alpha)$ we observe higher effective dimensions: it is approximately 27 for $\alpha=0.8$ and 19 for $\alpha=0.9$.  There is a qualitative difference between behaviour of $\bar{p}_y(\alpha)$ for empirical database and for the equidistribution. For the equidistribution on the sphere, $\bar{p}_y(\alpha)$ decreases approaching the point $\alpha=1$  from below like $const\times (1-\alpha)^{(n-1)/2}$. In the logarithmic coordinates it should look like $const+0.5(n-1)\ln(1-\alpha)$, exactly as we can see in Fig.~\ref{Fig:pwithoutlier}. The behaviour of $\bar{p}_y(\alpha)$ for  the LWF database is different and demonstrates some saturation rather than decay to zero. The human inspection of the inseparability at $\alpha=0.99$ shows that there are two pairs of images, which completely coincide but are labelled differently. This is an error in labelling. After fixing this error, the saturation vanished (Fig.~\ref{Fig:pwithoutlier}b) but the obvious difference between curves for the empirical data and the data for equidistribution on the sphere remained. We suppose that this difference is caused by the multicluster structure of the preprocessed LWF database.

\subsection{Fisher's separability for different probability distributions} 

First stochastic separation theorems about Fisher's separability  were proved for points randomly and independently sampled from an equidistribution in a ball \cite{GorbanTyuRom2016}. These results were extended later to the product distributions in a cube \cite{GorbTyu2017}.

Theorem \ref{Theorem:ExclVol2} and two following theorems  \cite{GorbTyu2017} demonstrate that Fisher's  discriminants are powerful in high dimensions.

\begin{theorem}[Equidistribution in $\mathbb{B}_n$]\label{ball1point}
Let $\{\boldsymbol{x}_1, \ldots , \boldsymbol{x}_M\}$ be a set of $M$  i.i.d. random points sampled  from the equidistribution in the unit ball $\mathbb{B}_n$. Let $0<r<1$ and $\rho=\sqrt{1-r^2}$. Then
\begin{equation}
\begin{split}\label{Eq:ball1}
\mathbf{P}&\left(\|\boldsymbol{x}_M\|>r \mbox{ and }
\left(\boldsymbol{x}_i,\frac{\boldsymbol{x}_M}{\| \boldsymbol{x}_M\| } \right)
<r \mbox{ for all } i\neq M \right) \\& \geq 1-r^n-0.5(M-1) \rho^{n};
\end{split}
\end{equation}
\begin{equation}
\begin{split}\label{Eq:ballM}
\mathbf{P}&\left(\|\boldsymbol{x}_j\|>r  \mbox{ and } \left(\boldsymbol{x}_i,\frac{\boldsymbol{x}_j}{\| \boldsymbol{x}_j\|}\right)<r \mbox{ for all } i,j, \, i\neq j\right) \\& \geq  1-Mr^n-0.5M(M-1)\rho^{n};
\end{split}
\end{equation}
\begin{equation}
\begin{split}\label{Eq:ballMangle}
\mathbf{P}&\left(\|\boldsymbol{x}_j\|>r  \mbox{ and } \left(\frac{\boldsymbol{x}_i}{\| \boldsymbol{x}_i\|},\frac{\boldsymbol{x}_j}{\| \boldsymbol{x}_j\|}\right)<r \mbox{ for all } i,j, \,i\neq j\right)\\&  \geq  1-Mr^n-M(M-1)\rho^{n}.
\end{split}
\end{equation}
\end{theorem}
Inequalities (\ref{Eq:ball1}), (\ref{Eq:ballM}), and (\ref{Eq:ballMangle}) are also closely related to  Proposition~\ref{Prop:ExclVol}.
According to Theorem \ref{ball1point}, the probability that a single element $\bfx_M$ from the sample $\mathcal{S}=\{\bfx_1,\dots,\bfx_{M}\}$ is linearly separated from the set $\mathcal{S}\setminus \{\bfx_M\}$ by the hyperplane $l(x)=r$ is at least
\[
1-r^n-0.5(M-1)\left(1-r^2\right)^{\frac{n}{2}}.
\]
This probability estimate depends on both $M=|\mathcal{S}|$ and dimensionality $n$.  An interesting consequence of the theorem is that if one picks a probability value, say $1-\vartheta$, then the maximal possible values of $M$ for which the set $\mathcal{S}$ remains linearly separable with  probability that is no less than $1-\vartheta$ grows at least exponentially with $n$. In particular, the following corollary holds

 \begin{corollary}\label{cor:exponential}
 Let $\{\boldsymbol{x}_1, \ldots , \boldsymbol{x}_M\}$ be a set of $M$ i.i.d. random points  from the equidistribution in the unit ball $\mathbb{B}_n$. Let $0<r,\vartheta<1$, and $\rho=\sqrt{1-r^2}$. If
\begin{equation}\label{EstimateMball}
M<2({\vartheta-r^n})/{\rho^{n}},
 \end{equation}
 then
 $
 \mathbf{P}((\boldsymbol{x}_i,\boldsymbol{x}_M{)}<r\|\boldsymbol{x}_M\| \mbox{ for all } i=1,\ldots, M-1)>1-\vartheta.
$
 If
 \begin{equation}\label{EstimateM2ball}
M<({r}/{\rho})^n\left(-1+\sqrt{1+{2 \vartheta \rho^n}/{r^{2n}}}\right),
 \end{equation}
  then $\mathbf{P}((\boldsymbol{x}_i,\boldsymbol{x}_j)<r\|\boldsymbol{x}_i\| \mbox{ for all } i,j=1,\ldots, M, \, i\neq j)\geq 1-\vartheta.$

  In particular, if inequality (\ref{EstimateM2ball}) holds then the set $\{\boldsymbol{x}_1, \ldots , \boldsymbol{x}_M\}$ is  Fisher-separable  with probability $p>1-\vartheta$.
 \end{corollary}

{Note that (\ref{Eq:ballMangle}) implies that elements of the set $\{\boldsymbol{x}_1, \ldots , \boldsymbol{x}_M\}$ are pair-wise almost or $\varepsilon$-orthogonal, i.e. $|\cos(\bfx_i,\bfx_j)|\leq \varepsilon$ for   $\varepsilon=r$ and all $i\neq j$, $1\leq i,j\leq M$,  with probability larger or equal than  $1-2Mr^n-2M(M-1)\rho^{n}$. Similar to Corollary \ref{cor:exponential}, one  can conclude that the cardinality $M$ of samples with such properties grows at least exponentially with $n$. The existence of the phenomenon has been demonstrated in \cite{Kurkova1993}. Theorem \ref{ball1point},  Eq. (\ref{Eq:ballMangle}), shows that the phenomenon is typical in some sense (cf.  \cite{GorbTyuProSof2016}, \cite{Kurkova:2017}).}

The linear separability property of finite but exponentially large samples of random i.i.d. elements is not restricted to equidistributions in a ball $\mathbb{B}_n$. As has been noted in  \cite{GorbanRomBurtTyu2016}, it holds for equidistributions in ellipsoids as well as for the Gaussian distributions. Moreover, it was generalized to product distributions in a unit cube. Consider, e.g. {the case}  when coordinates of the vectors $\bfx=(X_1,\dots,X_n)$ in the set $\mathcal{S}$ are independent random variables $X_i$, $i=1,\dots,n$ with expectations $\overline{X}_i$ and variances $\sigma_i^2>\sigma_0^2>0$. Let $0\leq X_i\leq 1$ for all $i=1,\dots,n$. The following analogue of Theorem \ref{ball1point} can now be stated.

\begin{theorem}[Product distribution in a cube \cite{GorbTyu2017}]\label{cube} Let  $\{\boldsymbol{x}_1, \ldots , \boldsymbol{x}_M\}$ be i.i.d. random points from the product distribution in a unit cube. Let
\[
R_0^2=\sum_i \sigma_i^2\geq n\sigma_0^2.
\]
Assume that the data are centralised and  $0< \delta <2/3$. Then
\begin{equation}\label{Eq:cube1}
\begin{split}
\mathbf{P}&\left(1-\delta  \leq \frac{\|\boldsymbol{x}_j\|^2}{R^2_0}\leq 1+\delta \mbox{ and }
\frac{(\boldsymbol{x}_i,\boldsymbol{x}_M)}{R_0\| \boldsymbol{x}_M\| }<\sqrt{1-\delta}   \mbox{ for all } i,j, \, i\neq M \right) \\ &\geq 1- 2M\exp \left(-2\delta^2 R_0^4/n \right) -(M-1)\exp \left(-2R_0^4(2-3 \delta)^2/n\right);
\end{split}
\end{equation}
\begin{equation}\label{Eq:cube2}
\begin{split}
\mathbf{P}&\left(1-\delta  \leq \frac{\|\boldsymbol{x}_j\|^2}{R^2_0}\leq 1+\delta \mbox{ and }
\frac{(\boldsymbol{x}_i,\boldsymbol{x}_j)}{R_0\| \boldsymbol{x}_j\| }<\sqrt{1-\delta} \mbox{ for all } i,j, \, i\neq j \right) \\& \geq 1- 2M\exp \left(-2\delta^2 R_0^4/n \right) -M(M-1)\exp \left(-2R_0^4(2-3 \delta)^2/n\right).
\end{split}
\end{equation}
\end{theorem}

In particular, under the conditions of Theorem \ref{cube}, set $\{\boldsymbol{x}_1, \ldots , \boldsymbol{x}_M\}$ is Fisher-separable  with probability $p>1-\vartheta$, provided that $M \leq ab^n$, where $a>0$ and $b>1$ are some constants depending only on $\vartheta$ and $\sigma_0$.

The proof  of Theorem \ref{cube} is based on concentration inequalities in product spaces \cite{Talagrand1995}. Numerous generalisations of Theorems \ref{ball1point} and \ref{cube} are possible for different classes of distributions, for example, for weakly dependent variables, etc.

We can see from Theorem \ref{cube} that the discriminant (\ref{discriminant}) works without precise whitening. Just the absence of strong degeneration is required: the support of the distribution contains in the unit cube (that is bounding from above) and, at the same time, the variance of each coordinate is bounded from below by $\sigma_0>0$.

Various generalisations of these theorems were proved in \cite{GorbanGolubGrechTyu2018}, for example, for log-concave distributions. The distribution  with density $\rho(\boldsymbol{x})$ is {\em log-concave} if set $D=\{\boldsymbol{x} | \rho(\boldsymbol{x})>0\}$ is convex and $g(\boldsymbol{x}) =-\log \rho(\boldsymbol{x})$ is convex function on  $D$. It is {\em strongly log-concave} if there exists a constant $c>0$ such that 
\begin{equation}\label{s-long-concave-def}
\frac{g(\boldsymbol{x})+g(\boldsymbol{y})}{2} - g\left(\frac{\boldsymbol{x}+\boldsymbol{y}}{2}\right) \geq c \|\boldsymbol{x}-\boldsymbol{y}\|^2, \quad\quad \forall \boldsymbol{x},\boldsymbol{y} \in D.
\end{equation}
The distribution is called {\em isotropic} if it is centralised and  its covariance matrix is identity matrix.
\begin{theorem}\label{log-con}
Let $S=\{\boldsymbol{x}_1, \ldots , \boldsymbol{x}_M\}$ be a set of $M$ i.i.d. random points sampled  from an isotropic log-concave distribution in ${\mathbb R}^n$. Then set $S$ is Fisher-separable with probability greater than $p=1-\psi$, $1>\psi>0$, provided that
$$
M \leq a \gamma^{\sqrt{n}},
$$
where $a>0$ and $\gamma>1$ are constants, depending only on $\psi$.
\end{theorem}
For strongly log-concave distributions (\ref{s-long-concave-def}) we return to the exponential estimates of $M$.
\begin{theorem}\label{st-log-con}
Let $S=\{\boldsymbol{x}_1, \ldots , \boldsymbol{x}_M\}$ be a set of $M$ i.i.d. random points sampled  from an isotropic strongly  log-concave distribution in ${\mathbb R}^n$. Then set $S$ is Fisher-separable with probability greater than $p=1-\psi$, $1>\psi>0$, provided that
$$
M \leq a \sqrt{\psi} \eta^n,
$$
where constants $a>0$ and $\eta>1$ depend only on $c$ from definition of strong long-concavity (\ref{s-long-concave-def}).
\end{theorem}
\begin{remark}\label{GaussLogConc}
Isotropic Gaussian distribution is strongly log-concave with $c=1/8$. Therefore,  Theorem~\ref{st-log-con} can be applied to Gaussian distribution too.
\end{remark}

Many other versions of the stochastic separation theorems can be proven. Qualitatively, all these results sound similarly:  exponentially large samples can be separated  with high probability by Fisher's discriminants for  essentially high-dimensional distributions without extremely heavy tails at infinity.

\subsection{Linear separation and SmAC distributions}
						
Linear separation is more general property than separation by Fisher's discriminant: it requires existence of a  linear functional that separates a point from a set. Two  approaches to construction linear separation are standard in AI applications: the historically first Rosenblatt perceptron algorithm and the Support Vector Machine (SVM). Of course, Fisher's discriminant is robust and much simpler in computations but one can expect from SVM more separation ability.  

General linear separability was analysed by Donoho and Tanner \cite{DonohoTanner2009}. They studied $m$-element i.i.d. samples of standard normal distribution in dimension $n$ and noticed that
if the ratio $n/m$ is fixed and dimension is large then all of the
points are on the boundary of the convex hull (of course, this is true even for exponentially large sample sizes $m$  and for the stronger property, Fisher's separability, see our Theorem~\ref{st-log-con} and Remark~\ref{GaussLogConc}). This observation contradicts the intuition. They have studied  various distributions numerically, observed the same effects for several highly non-i.i.d. ensembles too, and formulated the open problem: `Characterize the universality class containing the standard Gaussian: i.e. the class of matrix ensembles leading to phase transitions matching those for Gaussian polytopes.' 

The standard approach  {assumes that} the random set consists of independent identically distributed (i.i.d.) random vectors. The stochastic separation theorem presented below does not assume that the points are identically distributed. It can be very important: in the real practice the new data points are not compulsory taken from the same distribution as the previous points. In this sense the typical situation with the real data flow is far from an i.i.d. sample (we are grateful to G. Hinton for this important remark). New Theorem~\ref{th:separation} (below) gives also an answer to the {\em open problem} \cite{DonohoTanner2009}: it gives the general characterisation of the wide class of distributions with stochastic separation theorems (the SmAC condition below). Roughly speaking, this class consists of distributions without sharp peaks in sets with exponentially small volume (the precise formulation is below). We call this property ``SMeared Absolute Continuity'' (or SmAC for short) with respect to the Lebesgue measure: the absolute continuity means that the sets of zero measure have zero probability, and the SmAC condition  below requires that the sets with exponentially small volume should not have high probability.

Consider a  \emph{family} of distributions, one for each pair of positive integers $M$ and $n$. The general SmAC condition is
\begin{definition}\label{Def:SmAC}
The joint distribution of $\boldsymbol{x}_1, \boldsymbol{ x}_2, \dots, \boldsymbol{ x}_M$ has SmAC property if there exist constants $A>0$, {$B\in(0,1)$}, and $C>0$, such that for every positive integer $n$, any convex set $S \in {\mathbb R}^n$ such that
$$
\frac{V_n(S)}{V_n({\mathbb{B}}_n)} \leq A^n,
$$
any index $i\in\{1,2,\dots,M\}$, and any points $\boldsymbol{ y}_1, \dots, \boldsymbol{ y}_{i-1}, \boldsymbol{ y}_{i+1}, \dots, \boldsymbol{ y}_M$ in ${\mathbb R}^n$,
we have
{\begin{equation}\label{eq:condstar}
{\mathbf{P}}(\boldsymbol{ x}_i \in {\mathbb{B}}_n \setminus S\, | \, \boldsymbol{ x}_j=\boldsymbol{ y}_j, \forall j \neq i) \geq 1-CB^n.
\end{equation}}
\end{definition}

 We remark that
\begin{itemize}
\item We do not require for SmAC condition to hold for \emph{all} $A<1$, just for \emph{some} $A>0$. However, constants $A$, $B$, and $C$ should be independent from $M$ and $n$.
\item  We do not require that $\boldsymbol{ x}_i$ are independent. If they are, \eqref{eq:condstar} simplifies to
$$
{\mathbf{ P}}(\boldsymbol{ x}_i \in {\mathbb{B}}_n \setminus S) \geq 1-CB^n.
$$
\item  We do not require that $\boldsymbol{ x}_i$ are identically distributed.
\item  The unit ball ${\mathbb{B}}_n$ in  SmAC condition can be replaced by an arbitrary ball, due to rescaling.
\item  We do not require the distribution to have a bounded support - points $\boldsymbol{ x}_i$ are allowed to be outside the ball, but with exponentially small probability.
\end{itemize}

The following proposition establishes a sufficient condition for SmAC condition to hold.

\begin{proposition}
Assume that $\boldsymbol{ x}_1, \boldsymbol{ x}_2, \dots, \boldsymbol{ x}_M$ are continuously distributed in ${\mathbb{B}}_n$ with conditional density satisfying
\begin{equation}\label{eq:condden}
\rho_n(\boldsymbol{ x}_i \,|\,\boldsymbol{ x}_j=\boldsymbol{ y}_j, \forall j \neq i) \leq \frac{C}{r^n V_n({\mathbb{B}}_n)}
\end{equation}
for any $n$, any index $i\in\{1,2,\dots,M\}$, and any points $\boldsymbol{ y}_1, \dots, \boldsymbol{ y}_{i-1}, \boldsymbol{ y}_{i+1}, \dots, \boldsymbol{ y}_M$ in ${\mathbb R}^n$, where $C>0$ and $r>0$ are some constants. Then SmAC condition holds with the same $C$, any $B \in (0,1)$, and $A=Br$.
\end{proposition}

If $\boldsymbol{ x}_1, \boldsymbol{ x}_2, \dots, \boldsymbol{ x}_M$ are independent with $\boldsymbol{ x}_i$ having density $\rho_{i,n}$ in $n$-dimensional unit ball ${\mathbb{B}}$, then \eqref{eq:condden} simplifies to
\begin{equation}\label{eq:indden}
\rho_{i,n}(\boldsymbol{ x}) \leq \frac{C}{r^n V_n({\mathbb{B}}_n)}, \quad \forall n, \, \forall i, \, \forall \boldsymbol{x} \in {\mathbb{B}}_n,
\end{equation}
where $C>0$ and $r>0$ are some constants.

With $r=1$, \eqref{eq:indden} implies that SmAC condition holds for probability distributions whose density is bounded by a constant times density $\rho_n^{uni}:=\frac{1}{V_n({\mathbb{B}}_n)}$ of uniform distribution in the unit ball. With arbitrary $r>0$, \eqref{eq:indden} implies that SmAC condition holds whenever ration $\rho_{i,n}/\rho_n^{uni}$ grows at most exponentially in $n$. This condition is general enough to hold for many distributions of practical interest.

\begin{example}{(Unit ball)}\label{ex:ball}
If $\boldsymbol{ x}_1, \boldsymbol{ x}_2, \dots, \boldsymbol{ x}_M$ are i.i.d. random points from the equidistribution in the unit ball, then \eqref{eq:indden} holds with $C=r=1$.
\end{example}
\begin{example}{({Randomly perturbed} data)}\label{ex:noisy}
Fix parameter $\epsilon \in (0,1)$ ({random perturbation} parameter).
Let $\boldsymbol{ y}_1, \boldsymbol{ y}_2, \dots, \boldsymbol{ y}_M$ be the set of $M$ arbitrary (not compulsory random) points inside the ball with radius $1-\epsilon$ in ${\mathbb R}^n$. They might be clustered in arbitrary way or  belong to a subspace of  low dimension, etc. Let
$\boldsymbol{ x}_i, i=1,2,\dots,M$ for each $i$ be points, selected uniformly at random from a ball with centre $\boldsymbol{ y}_i$ and radius $\epsilon$. We consider $\boldsymbol{ x}_i$ as a ``perturbed'' version of $\boldsymbol{ y}_i$.
Then \eqref{eq:indden} holds with $C=1$, $r=\epsilon$.
\end{example}

\begin{example}{(Uniform distribution in a cube)}\label{ex:cube}
Let $\boldsymbol{ x}_1, \boldsymbol{ x}_2, \dots, \boldsymbol{ x}_M$ be i.i.d. random points from the equidistribution in the unit cube. Without loss of generality, we can scale the cube to have side length $s=\sqrt{4/n}$. Then \eqref{eq:indden} holds with $r<\sqrt{\frac{2}{\pi e}}$. 
\end{example}
\begin{remark}
For the uniform distribution in a cube
\begin{equation*}
\begin{split}
&V_n({\mathbb{B}}_n)\rho_{i,n}(\boldsymbol{ x}) = \frac{V_n({\mathbb{B}}_n)}{(\sqrt{4/n})^n} = \frac{\pi^{n/2}/\Gamma(n/2+1)}{(4/n)^{n/2}}\\& < \frac{(\pi/4)^{n/2}n^{n/2}}{\Gamma(n/2)} \approx \frac{(\pi/4)^{n/2}n^{n/2}}{\sqrt{4\pi/n}(n/2e)^{n/2}} \leq \frac{1}{2\sqrt{\pi}}\left(\sqrt{\frac{\pi e}{2}}\right)^n,
\end{split}
\end{equation*}
where $\approx$ means Stirling's approximation for gamma function $\Gamma$.
\end{remark}

\begin{theorem}\label{th:separation}
Let $\{\boldsymbol{x}_1, \ldots , \boldsymbol{x}_M\}$ be a set of i.i.d.  random points in ${\mathbb R}^n$ from distribution satisfying SmAC condition. Then $\{\boldsymbol{x}_1, \ldots , \boldsymbol{x}_M\}$ is linearly separable with probability greater than $1-\psi$, $\psi>0$, provided that
$$
M \leq a b^n,
$$
where constant $a$ depends on $A$, $B$, and $\psi$, and $b$ depends on $A$ and $B$
\end{theorem}

For more technical details, proofs, and estimation of constants we refer to works \cite{GorbTyu2017,GorbTyuProSof2016,GorbanGrechukTykin2018,GorbanGolubGrechTyu2018,Sidorov2018}.

\subsection{Correctors of AI legacy systems: implementation and testing\label{Sec:Correctors}}

Examples illustrating the theory have been reported in several publications and case studies \cite{GorbanRomBurtTyu2016, TyuGor2017knowlege}. We discuss some of those results concerning the problem of correcting  legacy AI systems, including Convolutional Neural Networks (CNN) trained to detect objects in images. Networks of the latter class show remarkable performance on benchmarks and, reportedly, may outperform humans on popular classification tasks such as ILSVRC \cite{He:2016}. Nevertheless, as experience confirms, even the most recent state-of-the-art CNNs  eventually produce an inconsistent behaviour. Success in testing of AIs in the laboratories does not always lead to appropriate functioning  in realistic operational conditions. 

One of the most pronounced examples of such malfunctioning was published recently \cite{Foxx2018}. British police's facial recognition matches (`positives') are reported highly inaccurate (more than 90\% false positive). The experts in statistics mentioned in the further discussion that `figures showing inaccuracies of 98\% and 91\% are likely to be a misunderstanding of the statistics and are not verifiable' \cite{face2018}.  Despite of the details of this discussion, the large number of false positive recognitions of `criminals' leads to serious concerns about security of AI use.

Re-training of large legacy AI systems requires significant computational and informational resources. On the other hand, as theory in the previous sections suggests, these errors can be mitigated  by constructing cascades of linear hyperplanes.  Importantly, construction of these hyperplanes can be achieved by mere Fisher linear discriminants and hence the computational complexity scales linearly with the number of samples in the data as opposed to e.g. Support Vector Machines  whose worst-case computational complexity scales as a cube of the size of the data sample.

\subsubsection{Single-neuron error correction of a legacy AI system}

In \cite{GorbanRomBurtTyu2016} we considered a legacy AI system that was designed to detect images of pedestrians in a video frame. A central element of this system was a VGG-11 convolutional network \cite{Simonyan:2015}. Such networks   have simple homogeneous architecture and  at the same time exhibit  good classification capabilities in ImageNet competitions \cite{Russ:2015}.  The network was successfully trained on a set of  RGB images  comprising $114,000$  pedestrians (`positive'  class) and $375,000$  non-pedestrian (`negative'  class),   re-sized to $128\times128$ pixels.   

The network was then used to detect pedestrian in the NOTTINGHAM video \cite{Nottingham} consisting of $435$ frames taken with an action camera. The video contains $4039$ images of pedestrians (identified by human inspection), and neither of these images has been included in the training set. A multi-scale sliding window approach was used on each video frame to provide proposals that can be run through the trained network. These proposals were re-sized to $128\times128$ and passed through the network for classification. Non-maximum suppression is then applied to the positive proposals. Bounding boxes were drawn and we compared results to a ground truth and identified false positives.

The network was able to detect $2896$ of  the total $4039$ images of pedestrians but it also returned $189$ false positives. The task was then to check if filtering out some or all of these errors in the legacy AI's output could be achieved with the help of a single linear functional separating these errors from correct detections.

To construct this functional we had to define a suitable linear space and the corresponding representations of the legacy AI's state in this space. In this case, the AI's state was represented by the feature vectors $\bfx_i$  taken from  the second to last fully connected layer in our VGG-11 network.  These extracted feature vectors $\bfx_i$  had $4096$ real-valued components suggesting that the data could be  embedded in $\Real^{4096}$. For the purposes of constructing these single-funcional AI correctors we generated  $M=114, 000$ feature vectors $\bfx_i$ ($i=1,\dots, M$) from the positive images in the AI's own training set. These $M$ feature vectors have been combined into the set $\mathcal{M}$. In addition to the set $\mathcal{M}$ we generated the set $\mathcal{Y}$ comprised of the feature vectors $\bfx_j$ corresponding to the identified $189$ false positives.

Elements of the sets $\mathcal{M}$ and $\mathcal{Y}$ have been centred and projected onto the first $2000$ principal components of $\mathcal{M}$.  This produced pre-processed sets $\mathcal{M}'$ and $\mathcal{Y}'$. We then generated  varying-size  subsets $\mathcal{Y}_j'$ of $\mathcal{Y}'$ and constructed Fisher linear discriminants $(\cdot,\bfw_j)-c_j$ with the weight $\bfw_j$ and threshold $c_j$ for $\mathcal{M}'$, $\mathcal{Y}_j'$. The threshold $c_j$ was chosen so that $c_j=\min_{\bfx_i\in\mathcal{Y'_j}} (\bfx_i,\bfw_j)$.

Once the functional have been constructed and the threshold determined, the corresponding linear model was placed at the end of our detection pipeline. For any proposal given a positive score by the legacy AI network we extracted the   feature vector and then run it through the linear model (after subtraction of the mean of $\mathcal{M}$ and projection onto the first $2000$ principal components of the set $\mathcal{M}$). Any detection that gave a non-negative score  was consequently removed by turning its detection score negative.

\begin{figure*}[ht]
\centering
\includegraphics[width=0.45\textwidth]{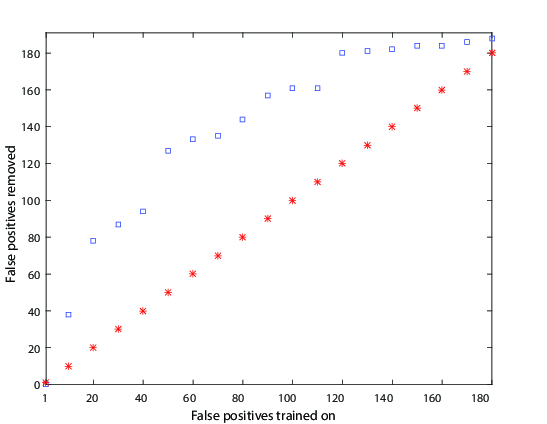} \hspace{3mm} \includegraphics[width=0.45\textwidth]{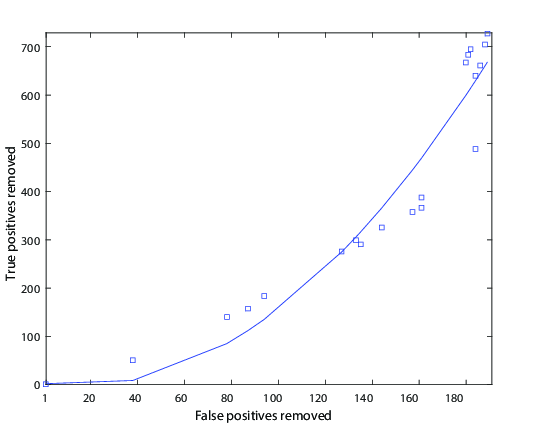}
\vspace{1mm}
\caption{Performance of {  {\it one-neuron} corrector built using} Fisher linear discriminant. {\it Left panel}: the number of false positives removed  as a function of the number of false positives the model was built on. Stars indicate the number of false positives used for building the model. Squares correspond to the number of false positives removed by the model. {\it Right panel}: the number of true positives removed as a function of the number of false positives removed. The actual measurements are shown as squares. Solid line is the least-square fit of a quadratic.}\label{fig:fp_fisher_effect}
\vspace{5mm}
\end{figure*}

Figure \ref{fig:fp_fisher_effect} shows the typical performance of the legacy AI system with the corrector.
Single false positive were removed without any detriment   on the true positive rates. Increasing the number of false positives that the single neuron model is to filter resulted in gradual deterioration of the true positive rate.

\begin{figure}
\centering
\includegraphics[width=0.46\textwidth]{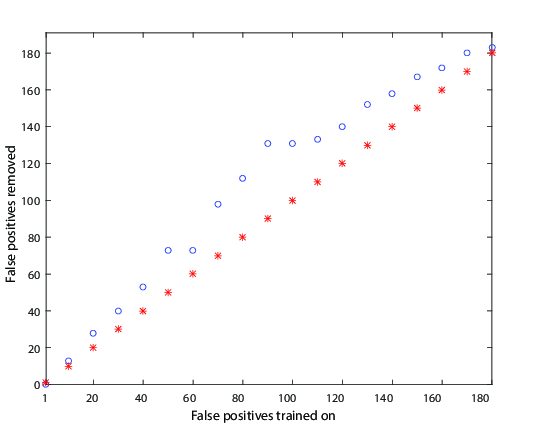} \hspace{3mm} \includegraphics[width=0.46\textwidth]{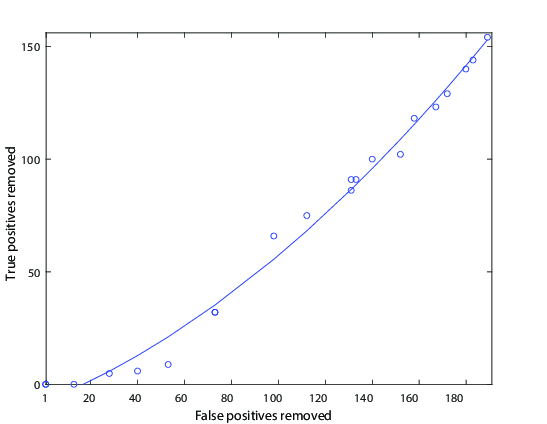}
\vspace{2mm}
\caption{Performance of { {\it one-neuron} corrector constructed using} the linear Support Vector Machine model. {\it Left panel}: the number of false positives removed as a function of the number of false positives the model was trained on. Stars indicate the number of false positives used for training the model. Circles correspond to the number of false positives removed by the model {\it Right panel}: the number of true positives removed as a function of the number of false positives removed. The actual measurements are presented as circles. Solid line is the least-square fit of a quadratic.}\label{fig:fp_svm_effect}
\end{figure}

In addition to linear model based on the Fisher discriminant, we have also used soft-margin  Support Vector Machines as a benchmark (Fig. \ref{fig:fp_svm_effect}). The true positives were taken from the CNN training data set.
The SVM model successfully removes $13$ false positives without affecting the true positives rate. Its performance deteriorates at a much slower rate than for the Fisher model. This, however, is balanced by the fact that the Fisher cap model removed significantly more false positives than it was trained on.

From the results shown in Figs. \ref{fig:fp_fisher_effect}, \ref{fig:fp_svm_effect}  it is evident that simple single-neuron correctors are  capable of correcting mistakes of legacy AI systems. Filtering large numbers of errors   comes at some cost to the detection of true positive, as expected. As we increase the number of false positives that we train our linear model on,  the number of removed true positives starts to rise too. Our tests illustrate, albeit empirically, that the separation theorems are indeed viable in applications. It is also evident that in this particular test  generalization capacity  of Fisher discriminant was significantly higher than that of the Support Vector Machines. As can be seen from Fig. \ref{fig:fp_fisher_effect}, Fig. \ref{fig:fp_svm_effect}, left panels, Fisher discriminants are more efficient in 'cutting off' larger number of mistakes than they have been trained on. In the operational domain near the origin in Fig. \ref{fig:fp_fisher_effect}, they do so without doing much damage to the legacy AI system. The Support Vector Machines performed better for larger values of false positives. This is not surprising as their training is iterative and accounts for much more information about the data than simply the sample covariance and mean.

\subsubsection{Knowledge transfer between AI systems}

In \cite{TyuGor2017knowlege} we made a step forward from  one-neuron error correction   and showed that the theoretical framework of Stochastic Separation Theorems can be employed to produce simple and computationally efficient algorithms for automated knowledge spreading between AI systems. Spreading of knowledge was achieved via small add-on networks constructed from   one-neuron correctors and supplementing decision-making circuits.

To illustrate the concept we considered two AI systems, a teacher  and a student. These two systems have been developed for the same task: to detect pedestrians  in live video streams. The teacher AI was modeled by an adapted  SqueezeNet \cite{iandola2016squeezenet} with about $725$K trainable parameters. The network was trained on a `teacher' dataset comprised of $554$K non-pedestrian (negatives), and $56$K pedestrian (positives) images. Pedestrian images have then been subjected to standard augmentation accounting for various geometric and colour perturbations.  The student AI was modelled by a linear classifier with Histograms of Oriented Gradients features \cite{Dalal:2005} and $2016$ trainable parameters. The values of these parameters were the result of the student AI training on a ``student'' dataset, a sub-sample of the ``teacher'' dataset comprising $16$K positives ($55$K after augmentation) and $130$K negatives, respectively. The choice of student and teacher AI  systems enabled us to emulate interaction between low-power edge-based AIs and their more powerful counterparts that could be deployed on a higher-spec embedded system or,  possibly, on a  server, or in  a computational cloud.

\begin{figure}
\centering
\includegraphics[width=0.4\textwidth]{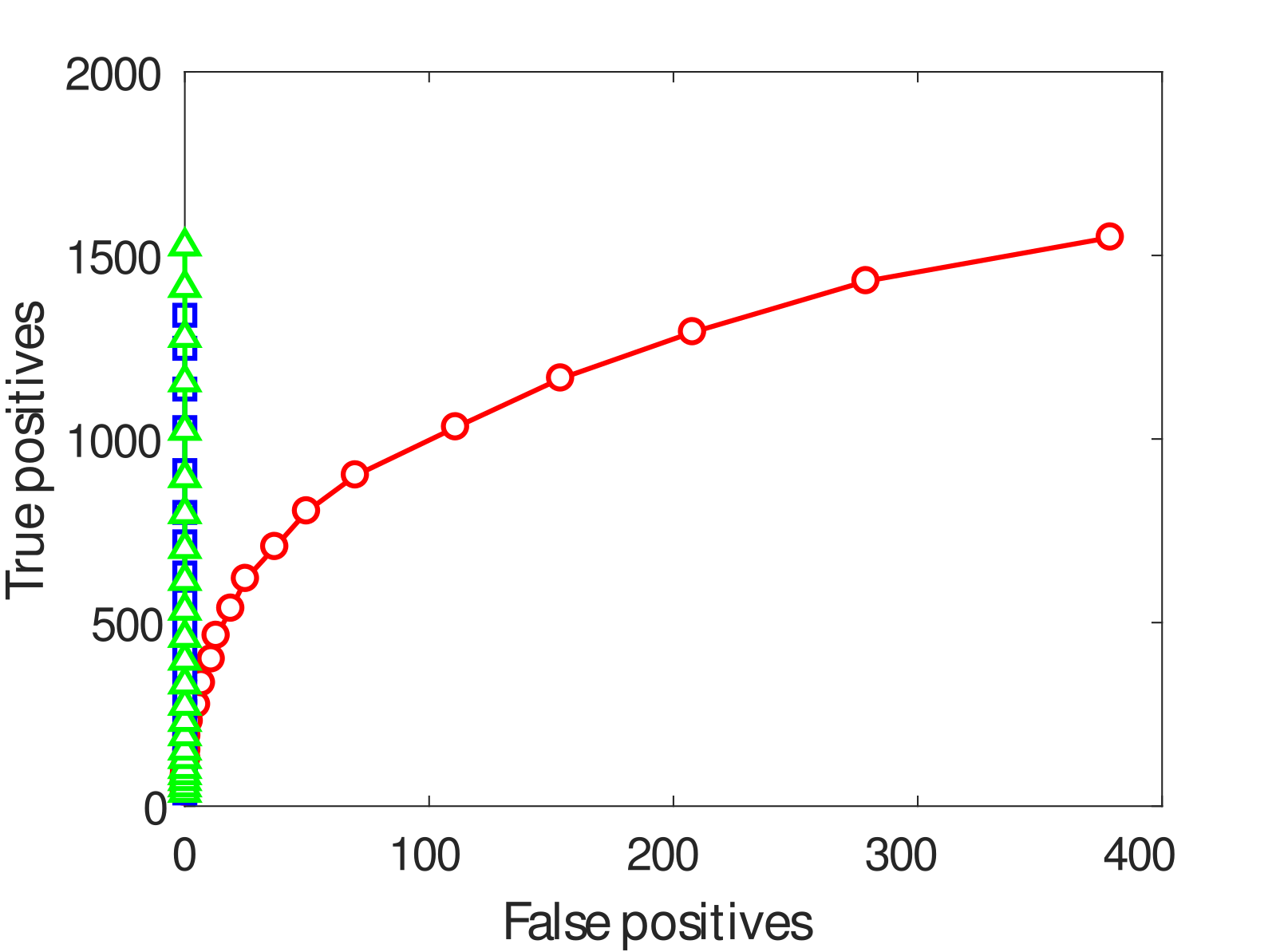} \hspace{5mm}\includegraphics[width=0.4\textwidth]{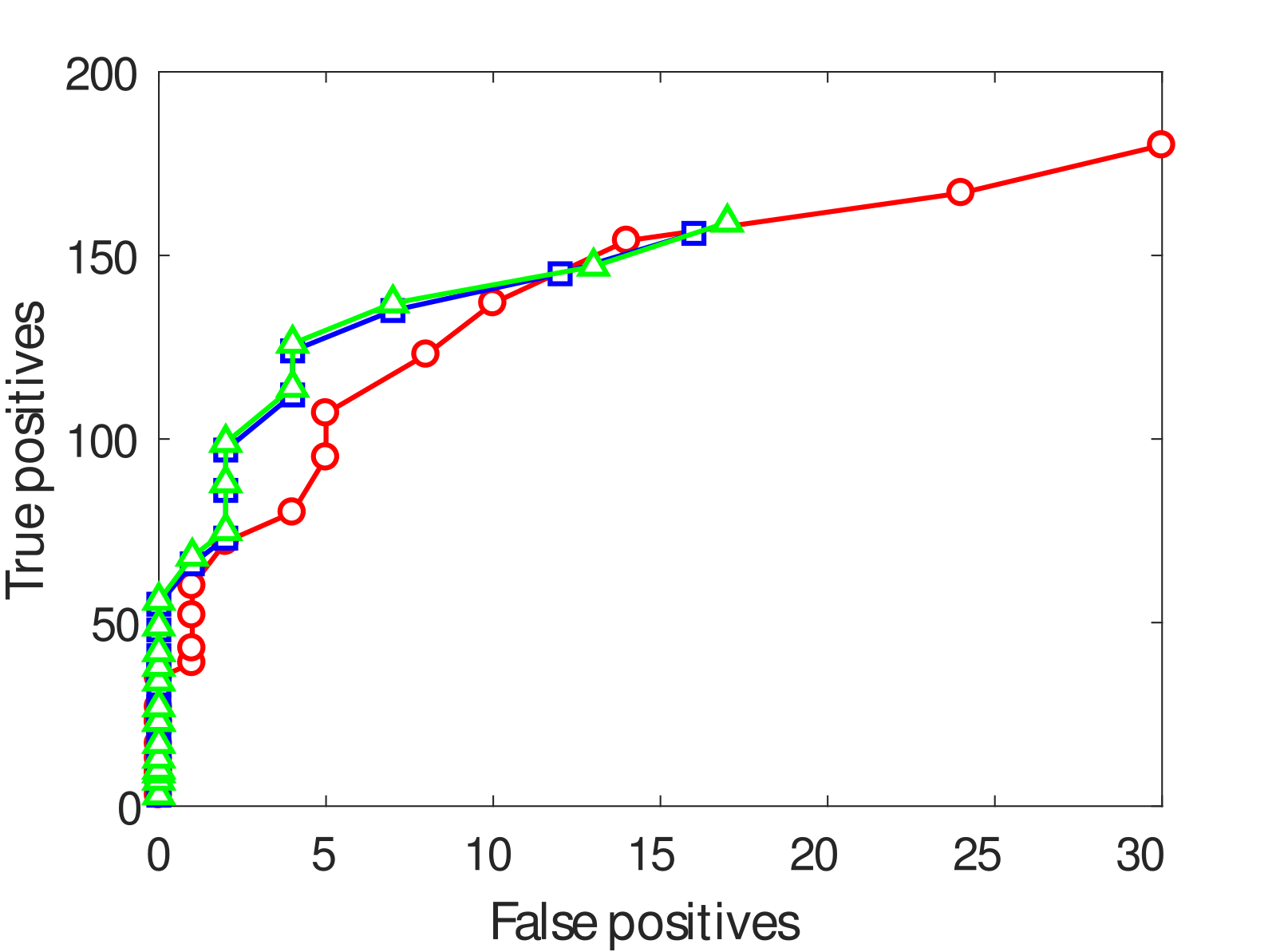}
\caption{Performance of the student AI on the training  set (left panel) and on the testing set (right panel). Red circles correspond to student AI without correcting networks. Blue squares capture performance of the student AI with a correcting network comprised of $30$ nodes, and green triangles correspond to the student AI supplemented with a $60$-node correcting network.}\label{fig:KT_example}
\end{figure}

The approach was tested on several video sequences, including the NOTTINGHAM video \citep{Nottingham} we used before. For each frame, the student AI system decided whether it contains an image of pedestrian. These proposals were then evaluated by the teacher AI. If a pedestrian shape was confirmed by the teacher AI in the reported proposal then no actions were taken. Otherwise an error (false positive) was reported, and the corresponding proposal along with its feature vector were retained for the purposes of uptraining and testing the student AI. The set of errors was then used to produce new training and testing sets for the student AI. This followed by construction of small networks  correcting errors of the student AI (see \cite{TyuGor2017knowlege}  for further details). It is worthwhile to note  that positives from the video were not included in the training set used to produce these networks. Performance of the student AI with and without correcting networks on  the NOTTINGHAM video   is illustrated with Fig. \ref{fig:KT_example}. As we can see from the figure, new knowledge (corrections of erroneous detects) have been successfully spread to the student AI. Moreover, the uptrained student system showed some degree of generalization of new skills which can be explained by correlations between feature vectors  corresponding to false positives.

 \section{Encoding and rapid learning of memories by single neurons\label{Sec:BMBLearning}}
 
 \subsection{The problems with codifying memories}
 
\begin{figure} 
\centering{\includegraphics[width = 0.95\textwidth]{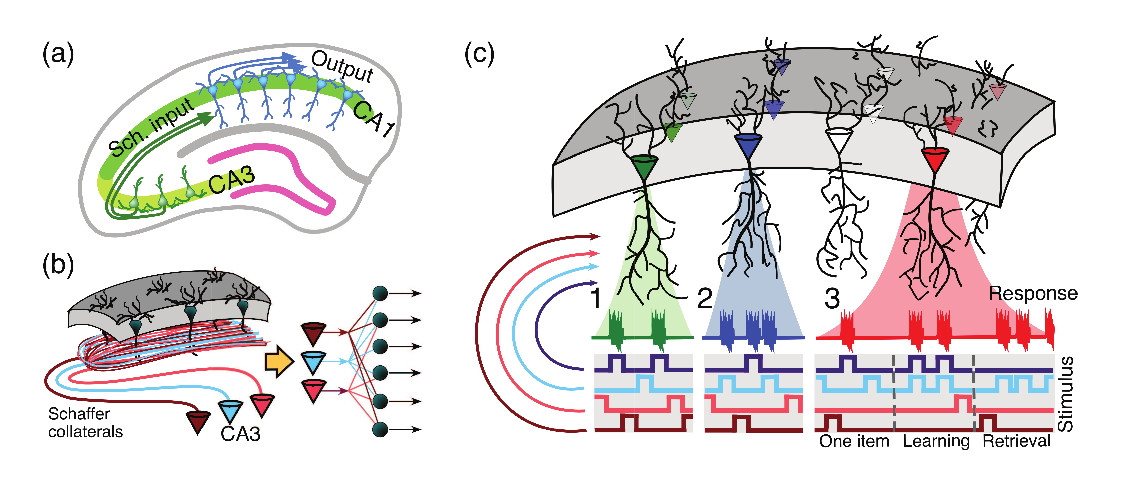}
}
\caption{Organisation of encoding memories by single neurons in laminar structures: (a) Laminar organization of the CA3 and CA1 areas in the hippocampus facilitates multiple parallel synaptic contacts between neurons in these areas by means of Schaffer collaterals; (b) Axons from CA3 pyramidal neurons bifurcate and pass through the CA1 area in parallel (left panel) giving rise to the convergence-divergence of the information content (right panel). Multiple CA1 neurons receive multiple synaptic contacts from CA3 neurons; (c) Schematic representation of three memory encoding schemes: (1) Selectivity. A neuron ({shown} in green) receives inputs from multiple presynaptic cells that code different information items. It detects (responds to) only one stimulus (purple trace), whereas rejecting the others; (2) Clustering. Similar to 1, but now a neuron ({shown} in blue) detects a group of stimuli (purple and blue traces) and ignores the others; (3) Acquiring memories. A neuron (shown in red) learns dynamically a new memory item (blue trace) by associating it with a known one (purple trace). }
\label{Fig1}
\end{figure}

Some of the memory functions are performed by stratified brain structures, e.g., the hippocampus.  The CA1 region of the hippocampus is constituted by a monolayer of morphologically similar pyramidal cells oriented with their main axis in parallel (Fig. \ref{Fig1}a). One of the major excitatory inputs to these neurons comes  from the CA3 region through Schaffer collaterals \cite{Amaral1989,Ishizuka1990,Wittner2007}, which can be considered as a hub routing information among various brain structures. Each CA3 pyramidal neuron sends an axon that bifurcates and leaves multiple collaterals in the CA1 with dominant parallel orientation (Fig. \ref{Fig1}b). This structural organisation allows multiple parallel axons conveying multidimensional `spatial' information from one area (CA3) simultaneously leave synaptic contacts on multiple neurons in another area (CA1). Thus, we have simultaneous convergence and divergence of the information content (Fig. \ref{Fig1}b, right).
 
Experimental findings suggest that there exist mechanisms for efficient increase of the selectivity  of individual  synaptic contacts and, in turn, codifying of memories:
\begin{itemize}
\item The multiple CA1 pyramidal cells distributed in the rostro-caudal direction are activated near-synchronously by assemblies of simultaneously firing CA3 pyramidal cells  \cite{Ishizuka1990,Li1994,Benito2014}.
\item Thus, an ensemble of single neurons in the CA1 can receive simultaneously the same synaptic input (Fig. \ref{Fig1}b, left). 
\item These neurons have different topology and functional connectivity \cite{Finnerty1983}, therefore, their response to the same input can be different. 
\item Experimental \textit{in-vivo} results show that long term potentiation can significantly increase the spike transfer rate in the CA3-CA1 pathway \cite{Fernandez2012}. 
\end{itemize}

We now can formulate the fundamental questions about the information encoding and formation of memories by  single neurons and their ensembles in laminated brain structures  \cite{TyukinBrain2017}:

\begin{enumerate}
\item {\it Selectivity: Detection of one stimulus from a set} (Fig. \ref{Fig1}c.1). Pick an arbitrary stimulus from a reasonably large set such that a single neuron from a neuronal ensemble detects this stimulus, i.e. generates a response. What is the probability that this neuron is stimulus-specific, i.e., it  rejects all the other stimuli from the set?
\item {\it Clustering: Detection of a group of stimuli from a set} (Fig. \ref{Fig1}c.2). Within a set of stimuli we select a smaller subset, i.e., a group of stimuli.  What is the probability that a neuron detecting all stimuli from this subset stays silent for all remaining stimuli in the set? Solution of this problem is expected to depend on the {\it similarity} of the stimuli inside the group and their dissimilarity from the remainder.
\item {\it Acquiring memories: Learning new stimulus by associating it with one already known} (Fig. \ref{Fig1}c.3). Let us consider two different stimuli $\boldsymbol{s}_1$ and $\boldsymbol{s}_2$ such that {for $t\leq t_0$} they do not overlap in time and a neuron detects  $\boldsymbol{s}_1$, but not $\boldsymbol{s}_2$. In the next interval $(t_0,t_1]$, {$t_1>t_0$} the stimuli start to overlap in time (i.e., they stimulate the neuron together). {For} $t> t_1$ the neuron receives only  stimulus $\boldsymbol{s}_2$.  What is the probability that for some $t_2\geq t_1$ the neuron detects $\boldsymbol{s}_2$?
\end{enumerate}
It may seem that for large sets of stimuli the probabilities in the stated questions should be small. Nevertheless, it is not the case and we aim to  show that they can be close to one even for exponentially large (with respect to dimension) sets of different stimuli.

Of course, evaluation of these probabilities depends on the `Outer World Models' and, therefore, cannot be exact. The standard machine learning assumptions \cite{Vapnik2000, cucker2002mathematical} are never correct but can bring some clarity before  more general and realistic models are employed. Following this standard, we assume that  the stimuli are generated in accordance with some distribution. Then, all stimuli that a neuron may receive are i.i.d. samples from this distribution.   Once a sample is generated, a stimuli sub-sample is independently selected for testing purposes. Relaxing these assumptions is possible (see theorems above). Nevertheless, they remain an important and useful first step of the analysis.

The outer world model should be supplemented by the model of a neuron and its adaptation. In this section we  show how the formulated three fundamental questions can be addressed within a simple  classical modelling framework, where  a neuron is represented by a  perceptron equipped with a Hebbian-type of learning. The efficiency of the mixture of learnable single neurons increases with dimension of data as it is expected in the light of stochastic separation theorems.

\subsection{The model: signals, neurons and learning}\label{sec:model_neuron}

We  follow conventional and rather general functional representation of signalling in the neuronal pathways. We assume that upon  receiving an input, a neuron can either generate a response or remain silent. Forms of the neuronal responses as well as the definitions of synaptic inputs vary from one model to another. Therefore, here we adopt a rather general functional approach. Under a stimulus we understand a number of excitations simultaneously (or within a short time window) arriving to a neuron through several axons (Fig.~\ref{Fig1}b) and thus transmitting some `spatially coded' information items \cite{Benito2016}. If the neuron responds to a stimulus (e.g., generates output spikes or increases its firing rate), we then say that the neuron \textit{detects} the informational content of the given  stimulus.

Consider a neuronal aggregate consisting of $L$ independent neurons. Each neuron receives a sequence of external stimuli through $n$ synaptic contacts. In what follows we will refer to $n$ as the `neuronal dimension'. Assume that there are $M$ different stimuli ($M$ can be large but finite), the $i$th individual stimulus  is modelled by a function 
\begin{equation}\label{eq:stimulus_definition}
\boldsymbol{s}_i(t)=\boldsymbol{x}_i \sum_{j}c(t-{\tau_{i,j}}), 
\end{equation}
where $\boldsymbol{x}_i $  is the $n$-dimensional vector of the stimulus content, its coordinates codify the information transmitted by $n$ individual axons, $\tau_{i,j}$ are the time moments of the $i$th stimulus presentation, the function $c(t)$ defines the time window of the stimulus context. For simplicity, we use a rectangular window:
\begin{equation}
\label{eq:spike_shape_definition}
c(t)=\left\{
\begin{array}{ll}
1, & \mbox{ if } t\in[0,\Delta T]  \\
0, & \mbox{ otherwise},
\end{array} \right .
\end{equation}
where $\Delta T > 0$ is the window width.

We assume that the frames of the presentation of {\em the same} stimulus do not overlap: for all $i$, $\tau_{i,j+1} > \tau_{i,j} + \Delta T$.

The overall input is the sum of the individual stimuli (\ref{eq:stimulus_definition}):
\begin{equation}\label{eq:NeurInput}
\boldsymbol{S}(t)  = \sum_{i,j} \boldsymbol{x}_i c(t-\tau_{i,j}).
\end{equation}

High dimension of the outer world is the crucial assumption: we analyse the {\em high-dimensional brain in a high-dimensional world}. Formally, this means that the distribution of the vectors $\boldsymbol{x}_i$ of the information content of stimuli is essentially high-dimensional. Various types of this assumption were discussed in Sec.~\ref{Sec:StochSep}: uniform distributions in $n$-dimensional  balls or cubes, Gaussian distributions, log-concave distributions, etc.

We use the simple classical {\em threshold model of neurons}. Each neuron has $n$ synaptic inputs. The $n$-dimensional vector of synaptic weights is $\boldsymbol{w}$. The membrane potential (the summarised synaptic input) of the neuron is the inner product 
$y=(\boldsymbol{w},\boldsymbol{S})$. The neuron generates a response $v=f(y-\theta)$, where $\theta$ is the firing threshold and $f$ is the transfer function; $f(z)=0$ for $z\leq 0$ and $f(z)>0$ for $z>0$.  In computational experiments we used   the `rump' activation function $f(z)=\max\{0,z\}$. Given a choice of the activation function $f$, the neuron is completely determined by the pair $(\boldsymbol{w}, \theta)$.

This threshold model disregards the specific dynamics of the neuronal response but even this simple model is already sufficient to clarify the fundamental problems of encoding memories \cite{TyukinBrain2017}. 

{\em Synaptic plasticity} is necessary for appearance of encoded memories. Dynamics of synaptic weights  $\boldsymbol{w}$ depends on the stimulus $\boldsymbol{S}$, the summarised synaptic input $y$, and the neuronal response $v$. According to a modified classical Oja rule \cite{Oja1982}, it is constructed as a  pursuit of the vector $\boldsymbol{S}/y$ by the synaptic weights vector $\boldsymbol{w}$:
\begin{equation}\label{eq:hebbian_oja}
\dot{\boldsymbol{w}}=\alpha  vy \left(\boldsymbol{S} -  \boldsymbol{w} y\right),\\
\end{equation}
with $\alpha=const>0$ and nonzero initial conditions
$\boldsymbol{w}(t_0)=\boldsymbol{w}_0 \neq 0.$

The difference from the classical Oja model is in the factor $v$: the rate of plasticity is proportional to the neuronal response. It nullifies for zero response: the pursuit of the stimuli content  by the synaptic weights is controlled by the neuronal response.

Solutions of the proposed model (\ref{eq:hebbian_oja})  remain bounded in forward time. Indeed,  after multiplication of (\ref{eq:hebbian_oja}) by $\boldsymbol{w}$ we obtain
\begin{equation}\label{normalisation}
\frac{1}{2}\frac{d \|\boldsymbol{w}\|^2}{dt}=\alpha v y^2(1-\|\boldsymbol{w}\|^2),
\end{equation}
where $\alpha>0$, $y^2\geq 0$, and $v\geq 0$ by construction. Therefore, if $\|\boldsymbol{w}\|\geq 1$ then $d(\|\boldsymbol{w}\|^2)/dt\leq 0$ and $\|\boldsymbol{w}\|$ does not grow.

\subsection{Selectivity 1: detection of one stimulus from a set}

All the stochastic separation theorems are, in their essence, theorems about selective detection of one stimulus from a large set (with rejection of all other stimuli from this set). For example, Proposition \ref{Prop:ExclVol} gives the following selectivity theorem (cf. \cite{TyukinBrain2017}).

\begin{theorem}\label{Theor:Select1}
Let $M$ $n$-dimensional vectors  $\boldsymbol{x}_i $ of the stimuli content belong to a ball in $\mathbb{R}^n$ with  centre $\boldsymbol{c}$ (mean stimulus) and radius $\rho$. Let $0<\psi <1$, $1/2<\alpha<1$,  and  $M<\psi (2\alpha)^n$. Assume that a new stimulus content $\boldsymbol{x}_{M+1}$ is sampled randomly from the equidistribution in this ball. Then with probability $p> 1-\psi$ the neuron $(\boldsymbol{w}, \theta)$ with 
\begin{equation}\label{wthetaseparation}
\boldsymbol{w}=\boldsymbol{x}_{M+1}-\boldsymbol{c}, \;\; \theta = \alpha (\boldsymbol{w},\boldsymbol{w})+ (\boldsymbol{w},\boldsymbol{c})
\end{equation}
detects selectively the stimulus with content $\boldsymbol{x}_{M+1}$ and rejects all the stimuli with content  $\boldsymbol{x}_{i}$ ($i=1, \ldots, M$).
\end{theorem}
Expression for $\theta$ in (\ref{wthetaseparation}) connects the neural threshold $\theta$ with the threshold $\alpha$ in Fisher's discriminant (\ref{discriminant}).

{\em Robustness of separation} is an important question: how wide is the interval of possible values of the synaptic weights for separation of the stimuli? The following extension of Theorem \ref{Theor:Select1} gives the answer to this question.

\begin{theorem}\label{Theor:Robust}
Let the conditions of Theorem \ref{Theor:Select1} hold,  $0<\xi<1-1/(2\alpha)$, and $M<\psi (1/2\alpha+\xi )^{-n}$. Then with probability $p>1-\psi$ every  neuron $(\boldsymbol{w'}, \theta)$ with
\begin{equation}\label{wthetaseparation1}
\frac{1}{\rho}\|\boldsymbol{w'}-\boldsymbol{w}\|<\xi, \;\;  \theta = \alpha (\boldsymbol{w'},\boldsymbol{w'})+ (\boldsymbol{w'},\boldsymbol{c})
\end{equation}
detects selectively the stimulus with content $\boldsymbol{x}_{M+1}$ and rejects all the stimuli with content  $\boldsymbol{x}_{i}$ ($i=1, \ldots, M$).
  \end{theorem}
To prove this theorem, it is sufficient to add a spheric layer of thickness $\xi/\rho$ to each  ball of excluded volume  (the coloured ball in Fig. \ref{Fig:Excluded}) and repeat the proof of Proposition \ref{Prop:ExclVol}. 

The parameter $\xi$ can be called a {\em  relative robustness}. It is the ratio of the possible allowed deviation $ \|\boldsymbol{w'}-\boldsymbol{w}\|$ to the radius  $\rho$ of the ball that includes the stimuli content vectors. According to Theorem \ref{Theor:Robust}, the allowed interval of  $\xi$ is $[0, 1-1/(2\alpha)]$. The upper bound of the number of stimuli $M$ decreases if $\xi$ grows, $M<\psi (1/2\alpha+\xi )^{-n}$ for given probability bound of wrong detection $\psi$, Fisher's threshold $\alpha$, and dimension of input signals $n$.

 Other estimates of the level of robustness for equidistribution of all stimuli content vectors in a ball were presented in \cite{TyukinBrain2017}.

The neuronal threshold $\theta$ in (\ref{wthetaseparation}), (\ref{wthetaseparation1}) is a sum of two terms. The first term, $\alpha (\boldsymbol{w},\boldsymbol{w})$  is just a right hand side of the definition of Fisher's  discriminant (\ref{discriminant}). 
The plasticity dynamics (\ref{eq:hebbian_oja}) produces normalised vectors of synaptic weights if the neuron is not silent at average  (i.e. the average value of the coefficient $v y^2$ in (\ref{normalisation}) does not vanish for large time). For normalised vectors  $\boldsymbol{w}$ the first term in expression for $\theta$ coincides with Fisher's discriminant threshold $\alpha$. 

 The second term, $ (\boldsymbol{w},\boldsymbol{c})$   appeared because the stimuli content vectors are not assumed to be centralised. Of course, for batch algorithms of machine learning, the input vectors are known in advance and the centralisation is a standard routine operation. This approach is also convenient for theoretical analysis. The situation is different for on-line learning: the stimuli are not known in advance and, moreover, there may be a `concept drift' and the sequence of stimuli may be very far from a stationary random process. Precise centralisation with constant centre $\boldsymbol{c}$ is impossible in such cases, but various moving average centralisation algorithms are applicable. In these methods, $\boldsymbol{c}$ depends on time and is evaluated as average in a time frame, average with exponential kernel or with linear combination of exponential kernels with different exponents, etc. Such methods are very popular for analysis of non-stationary time series, from financial analysis and econometrics to biometrics (see, for example, \cite{Brock1992}).

Recently, it was demonstrated {\em in vivo} that the `threshold adaptation' with various moving averages is an important part of adaptation of  neurons.  Slow voltage fluctuations do not contribute to spiking because they are filtered by threshold adaptation and the relevant time-dependent variable is not  the membrane potential, but rather its distance to a dynamic threshold, which is called the `effective signal' \cite{Fontaine2014}. The spike threshold  depends on the history of the membrane potential on many temporal scales and the threshold adaptation  allows neurons to  compute specifically and robustly and better detect well-correlated inputs \cite{Huang2016}.

These ideas lead to the cascade of  adaptation models. Instead of one model (\ref{eq:hebbian_oja}) we have  several pursuit models. For example, for two time scales (adaptation of average and adaptation of neurons) we can write two equations:

\begin{equation}\label{cascadeadapt}
\begin{split}
&\frac{d \boldsymbol{c}}{d t}=a (\boldsymbol{S}-\boldsymbol{c}); \\
&\frac{d \boldsymbol{w}}{d t}=\alpha v y (\boldsymbol{E}-\boldsymbol{w} y),
\end{split}
\end{equation}
where the rate of adaptation $a\geq 0$, $\boldsymbol{E}=\boldsymbol{S}-\boldsymbol{c}$ is the vector of `effective stimulus', $y=(\boldsymbol{w},\boldsymbol{E})$ is the `effective signal'  \cite{Fontaine2014},  and $v=f(y-\theta)$ is the neuronal response.

There are empirical evidences that the rate of average threshold adaptation $a$ depends on the neuronal response of a group of neurons and tends to zero if they remain silent \cite{Huang2016}. For example, we can assume $a=\alpha_c\sum v_i$, where $v_i$ are neuronal responses of the neurons from the group. 

Moreover, there may exist multiscale adaptation processes with different `averages' $ \boldsymbol{c}$: the long term average, the average deviation from the long term average during a shorter time, etc. Slower processes with smaller $\alpha_c$ can correspond to larger groups of neurons. Construction of such a cascade of equations is obvious. The last equation in this cascade is adaptation of an individual neuron, as  is written in (\ref{cascadeadapt}).

\subsection{Selectivity 2 (clustering): Detection of a group of stimuli from a set} 

For many models of the stimuli distribution, the possibility of selective detection  of stimuli from a cluster with rejection all other stimuli is determined by the separability of this cluster from the average stimuli content $\boldsymbol{c}$. Consider a set of `relevant' stimuli content vectors  $K$ that should be separated from all `irrelevant' stimuli. Assume that the stimuli content vectors are uniformly distributed in the $n$-dimensional ball with centre  $\boldsymbol{c}$ and radius $\rho$ and there exists such a normalised weight vector $\boldsymbol{w}$ and $h>0$ that
\begin{equation}\label{clusterSep}
(\boldsymbol{w},\boldsymbol{x}-\boldsymbol{c})>h
\end{equation}
for all $\boldsymbol{x}\in K$.

The cluster $K$ is contained in a half of the ball of radius $\rho_c=\sqrt{\rho^2-h^2}$ together with all vectors that satisfy the inequality (\ref{clusterSep}).
The probability $\psi_1$ that a randomly selected vector also belongs to this ball  can be evaluated as the ratio of volumes:
\begin{equation}\label{ClusterSeparationProb}
\psi_1<\frac{1}{2}\left(\frac{\rho_c}{\rho}\right)^n=\frac{1}{2}\left(1-\frac{h^2}{\rho^2}\right)^{n/2}\approx \frac{1}{2}\exp\left(-\frac{1}{2}\frac{h^2}{\rho^2}n\right)
\end{equation}
 This probability decays exponentially with $n$. The probability of selective rejection of a random stimulus is $p_1>1-\psi_1$. For $M$ stimuli the probability $\psi$ of at least one detection error does not exceed $M\psi_1$. With probability $p>1-M\psi_1$ there will be no errors for $M$ stimuli.  Therefore, for each fixed value of $p$ ($0<p<1$) we have an exponential estimate of possible number $M$ of different random stimuli, which can be separated from the cluster $K$ by the single neuron with  synaptic weights $\boldsymbol{w}$.
 
A uniform distribution in a high-dimensional ball is the simplest case for all estimates but at the same time it is not very far from the more general `essentially high-dimensional' distributions. For many such distributions the concentration theorems state  that they  are concentrated in a small vicinity of a spheric shell (ensemble equivalence).

It is natural to assume that the stimuli in a cluster are correlated. Nevertheless, the selective detection of a cluster is also possible for sets of independent stimuli. In high dimensions, i.i.d. vectors   randomly and independently sampled from an equidistribution in the unit ball (centred at the origin)  are almost orthonormal with high probability \cite{GorbTyuProSof2016}. Let  us assume that the number of vectors $k$ in cluster $K$ does not exceed $n$. For the sake of simplicity, we also assume that they are orthonormal (neglect deviations from orthogonality and impose unit norm). Then, according to elementary geometry (the Pythagoras theorem),   the normalised average vector of weights	
\begin{equation}\label{averageweight}
\boldsymbol{w}=\frac{\sum_{\boldsymbol{x}\in K} \boldsymbol{x}}{\left\|\sum_{\boldsymbol{x}\in K}  \boldsymbol{x}\right\|}
\end{equation}
 with high probability separates $K$ from the origin (\ref{clusterSep}) with $h=\gamma/\sqrt{k}$. Here $\gamma$ ($0<\gamma<1$) is a constant, which can be selected arbitrarily but should not change with $n$. The exponential estimates of probabilities of a single false positive detection $\psi_1$ (\ref{ClusterSeparationProb}) and of absence of false positives $p>1-M\psi_1$ also hold for this case: 
 $$\psi_1<\frac{1}{2}\left(1-\frac{\gamma^2}{k}\right)^{n/2}\approx \frac{1}{2}\exp\left(-\frac{\gamma^2}{2} \frac{n}{k}\right).$$
 Further  applications of quasiorthogonality to  assess  the of neuronal selectivity to multiple stimuli with detailed estimates of the probability of errors are presented in Section `Selectivity of a Single Neuron to Multiple Stimuli' of \cite{TyukinBrain2017}. We refer to this and the previous paper \cite{GorbTyuProSof2016} for the formal details and demonstrate below how these results refer to the ensemble equivalence and Maxwellian distribution. 
 
 An equidistribution in a unit high-dimensional ball is concentrated near its surface (`microcanonical ensemble', see Fig.~\ref{Volume}). Expectation of the vector length for large $n$ tends to 1: $E(\|\boldsymbol{x}\|^2)\approx 1$. At the same time, distribution of the coordinates of $\boldsymbol{x}$ converges to a Gaussian distribution with variance $1/n$ (Maxwellian distribution) and these coordinates are almost independent (`canonical ensemble'). For large $n$, equidistribution, microcanonical and canonical ensembles are almost equivalent: the corresponding average values of Lipschitz functions almost coincide. 
 
 Select $k$ independent vectors from the uniform distribution:  $K=\{\boldsymbol{x}_1, \ldots, \boldsymbol{x}_k\}$. We aim to evaluate the distribution of the inner product of $\boldsymbol{x}_i$ by the sum of these vectors. Because of independence, the choice of $i$ is not important and we set $i=1$. 
 \begin{equation}\label{productvectoronsum}
\left(\boldsymbol{x}_1, \sum_{\boldsymbol{x}\in K} \boldsymbol{x}\right)=\|\boldsymbol{x}_1\|\left(\|\boldsymbol{x}_1\|+\sum_{i=2}^k\left(\frac{\boldsymbol{x}_1}{\|\boldsymbol{x}_1\|},\boldsymbol{x}_j\right) \right)
 \end{equation}
 In this expression, $\|\boldsymbol{x}_1\|\approx 1$ (microcanonical ensemble) and the inner products $\left(\frac{\boldsymbol{x}_1}{\|\boldsymbol{x}_1\|},\boldsymbol{x}_j\right)$ can be considered just as  coordinates of  $\boldsymbol{x}_j$ (rotational invariance) and (almost) obey the Gaussian distribution with the variance $1/n$ (canonical ensemble, Maxwellian distribution). The sum of these coordinates also obeys the Gaussian distribution with variance $(k-1)/n$ (independence). The probability $\psi_-$ that the inner product (\ref{productvectoronsum}) is negative can be evaluated as the probability that random variable sampled from  the standard normal distribution with the mean  $\mu=0$ and the standard deviation $\sigma=1$ has the values above $\sqrt{n/k}$:
\begin{equation}\label{normal}
\psi_-\approx \frac{1}{2\pi}\int_{\sqrt{n/k}}^\infty \exp(-x^2/2)dx.
\end{equation}
 For $\sqrt{n/k}=2$ we get $\psi_-\approx 0,025$, and  $\psi_-\approx 0,0015$ for  $\sqrt{n/k}=3$. This $\psi_-$ is the probability that $\boldsymbol{x}_1$ is not separated from the origin by the   average vector of weights (\ref{averageweight}). Probability $p$ that all $\boldsymbol{x}\in K$ will be separated from the origin by  the   average vector of weights (\ref{averageweight}) can be evaluated as $p\geq 1-k\psi_-$. Modification of (\ref{normal}) for separation with a given gap $\theta$ is obvious.
 
Thus, if a cluster of relevant stimuli content is separable from the origin then it is also separable from an exponentially large set of random irrelevant stimuli.  A special and convenient case is separability by the weight vector, which coincides with the average content vector (\ref{averageweight}).   If the number of relevant stimuli content vectors is several times less  than neuronal dimension $n$, then the separation  by the average content vector is possible with high probability even for independent sampling of relevant vectors.

 \subsection{Acquiring memories: Learning new stimulus by associating it with one already known}
 
In this section we explicitly {account for} the time evolution of the synaptic weights, $ \boldsymbol{x}(t)$ in accordance with the modified Oja rule (\ref{eq:hebbian_oja}) and demonstrate how this evolution  gives rise to dynamic memories in single neurons.

We  deal with two sets of stimuli content, the relevant one, $\mathcal{Y}$, and the background one, $\mathcal{M}$. For probability evaluation, we assume that  $\mathcal{M}$ is an i.i.d. sample from the uniform distribution in the $n$-dimensional unit ball centred at the origin. $\mathcal{Y}$ is also a subset of this ball.

At the initial time moment, $t=0$, only one information relevant item $\boldsymbol{x}_0\in \mathcal{Y}$  is `known' to the neuron. The neuron rejects all other items, both  from $\mathcal{M}$ and $\mathcal{Y}$. Then a learning epoch is organised when all stimuli with content from the set $\mathcal{Y}$ arrive to a neuron completely synchronized. The overall input is the sum of the individual stimuli from  $\mathcal{Y}$: (\ref{eq:stimulus_definition}):
\begin{equation}\label{eq:NeurInputSinc}
\boldsymbol{S}(t)  = \sum_i \left(\sum_{\boldsymbol{x}\in \mathcal{Y}}  \boldsymbol{x}\right) c(t-\tau_{i}) \;\;\tau_{i+1} > \tau_{i} + \Delta T.
\end{equation}
Such a synchronization could be interpreted as  a mechanism for associating  different  information items for the purposes of memorizing them at a later stage.

The dynamics of the synaptic weights for $t>0$ is given Eq. (\ref{eq:hebbian_oja}) with the input signal (\ref{eq:NeurInputSinc}). Recall that the dynamics of norm $\|\boldsymbol{w}\|^2$ can be decoupled from the dynamics of $\boldsymbol{w}$ direction and $\|\boldsymbol{w}(t)\|\to 1$ when $t\to \infty$ if the integral
\begin{equation}\label{spiketime}
T_{\rm n.r.}(t)=\int_0^{t} v(\tau) y^2(\tau) d\tau
\end{equation}
diverges when $t\to \infty$. 

The `n.r. time' $T_{\rm n.r.}$ (n.r. stands for neuronal response) characterises time and intensity of neuronal response.

The unit sphere of the vectors of synaptic weights is stable invariant manifold of the dynamics of synaptic plasticity (\ref{eq:hebbian_oja}). Let us assume in the rest of this Section that $\|\boldsymbol{w}_0\|=1$. Then  $\|\boldsymbol{w}(t)\|= 1$ for $t>0$ and any stimuli $\boldsymbol{S}(t)$.

The trajectory of adaptation (\ref{eq:NeurInputSinc}) belongs to a circular arc (with the arc length $\beta<\pi$)  between $\boldsymbol{w}_0$ and the normalized average vector $$\bar{\boldsymbol{w}}=\frac{\sum_{\boldsymbol{x}\in \mathcal{Y}}\boldsymbol{x}}{ \|\sum_{\boldsymbol{x}\in \mathcal{Y}}\boldsymbol{x}\| }$$
if $\sum_{\boldsymbol{x}\in \mathcal{Y}}\boldsymbol{x}\neq 0$.
Each weight vector $\boldsymbol{w}$ on this arc is a linear combination $\boldsymbol{w}=a_0 \boldsymbol{w}_0+\bar{a}\bar{\boldsymbol{w}}$ with $0\leq a_0, \bar{a}\leq 1$ and $ a_0+ \bar{a}\geq 1$.

Let us formulate the simple conditions   that, during the learning phase the synaptic weights $\boldsymbol{w}(t)$ evolve in time {so} that the neuron becomes responsive to all $\boldsymbol{x} \in \mathcal{Y}$ whilst remaining silent to all $\boldsymbol{x} \in  \mathcal{M}$ (Fig. \ref{Fig1}c.3)?

The following conditions are obvious:
\begin{enumerate}
\item $(\boldsymbol{w}_0, \boldsymbol{x}_0)>\theta_0$;
\item $(\boldsymbol{w}_0, \boldsymbol{x})>-\frac{1}{|\mathcal{Y}|-1}(\theta_0-\theta)$  for all $\boldsymbol{x} \in \mathcal{Y}$, $\boldsymbol{x}\neq \boldsymbol{x}_0$;
\item $(\bar{\boldsymbol{w}},\boldsymbol{x})>\theta$ for all $\boldsymbol{x} \in \mathcal{Y}$;
\item $(\boldsymbol{w}_0, \boldsymbol{x})<\theta$ for all $\boldsymbol{x} \in  \mathcal{M}$;
 \item $(\bar{\boldsymbol{w}},\boldsymbol{x})<\theta$ for all $\boldsymbol{x} \in \mathcal{M}$.
\end{enumerate}
\begin{itemize}
\item The first condition means that the neuron $(\boldsymbol{w}_0, \theta_0)$ with a high initial threshold $\theta_0$ ($\theta_0>\theta$) detects  $\boldsymbol{x}_0$ .
\item According to the second condition,  the neuron $(\boldsymbol{w}_0, \theta)$ responds to the stimulus $\boldsymbol{S}$ with the content $\sum_{\boldsymbol{x}\in \mathcal{Y}}\boldsymbol{x}$ (\ref{eq:NeurInputSinc}).
\item Condition 3 means that the neuron $(\bar{\boldsymbol{w}}, \theta)$ responds to all the relevant stimuli with content from $\mathcal{Y}$.
\item According to conditions 4 and 5 both   neurons $(\boldsymbol{w}_0, \theta)$ and  $(\bar{\boldsymbol{w}}, \theta)$ remain silent to the irrelevant stimuli with content from $\mathcal{M}$.
\end{itemize}
Conditions 1-3 are sufficient to keep the neuron $(\boldsymbol{w}(t),\theta)$ responsive both  to the stimulus content $\boldsymbol{x}_0$ and  to the stimulus $\sum_{\boldsymbol{x}\in \mathcal{Y}}\boldsymbol{x}$ during adaptation (\ref{eq:NeurInputSinc}). Indeed,
 $$\boldsymbol{w}(t)=a_0(t) \boldsymbol{w}_0+\bar{a}(t)\bar{\boldsymbol{w}},$$ 
 $$(\boldsymbol{w}(t), \boldsymbol{x}_0)=a_0(t) (\boldsymbol{w}_0,\boldsymbol{x}_0)+\bar{a}(t)(\bar{\boldsymbol{w}},\boldsymbol{x}_0)>a_0(t) \theta_0 +\bar{a}(t)\theta \geq \theta,$$
$$ \mbox{and }\left(\boldsymbol{w}(t), \sum_{\boldsymbol{x}\in \mathcal{Y}}\boldsymbol{x}\right)>a_0(t) (\theta_0-(\theta_0-\theta))+\bar{a}(t)|\mathcal{Y}|\theta>\theta.$$  
  Therefore, due to the neural plasticity and the training dynamic equations  (\ref{eq:NeurInputSinc}) the synaptic weight vector converges from the initial condition $\boldsymbol{w}_0$ to the limit point $\bar{\boldsymbol{w}}$.
 
 Nevertheless, conditions 1-5 do not guarantee that in the transient states on the way from $\boldsymbol{w}_0$ to the limit point $\bar{\boldsymbol{w}}$ the neuron $(\boldsymbol{w}(t),\theta)$ will be silent to all the stimuli with content from $\mathcal{M}$. These temporary false positive errors in the transient training states may be a deep universal property of neural plasticity and associative memory. The effect increases with growth the angle $\angle(\boldsymbol{w}_0 ,\bar{\boldsymbol{w}})$.  If we need to  avoid such temporary false positives, then conditions 3 and 5 should be modified. An elementary geometric consideration gives the sufficient conditions:
\begin{enumerate}
\item[4'.]   $(\boldsymbol{w}_0, \boldsymbol{x})<\gamma \theta$ for all $\boldsymbol{x} \in  \mathcal{M}$;
\item[5'.]   $(\bar{\boldsymbol{w}},\boldsymbol{x})<\gamma \theta$ for all $\boldsymbol{x} \in \mathcal{M}$,
\end{enumerate}
where $$\gamma=\cos\frac{\angle(\boldsymbol{w}_0 ,\bar{\boldsymbol{w}})}{2}=\sqrt{\frac{1+(\boldsymbol{w}_0,\bar{\boldsymbol{w}})}{2}}<1.$$
Conditions 4' and 5' mean that the set of  irrelevant content $\mathcal{M}$ is separated from the relevant  content `with a gap'.

Let us evaluate the probability of the conditions for randomly selected stimuli content. Assume that the set of irrelevant stimuli content vectors  $\mathcal{M}$ is a random i.i.d. sample from the equidistribution in the unit $n$-dimensional ball. Probability $\psi_1$ that one random vector $\boldsymbol{x}$ violates condition 4 is estimated by the inequality (\ref{ClusterSeparationProb}) (with $\rho=1$ and $h=\theta$): $\psi_1<\frac{1}{2}(1-\theta^2)^{n/2}\approx \frac{1}{2}\exp(-\frac{1}{2}\theta^2 n)$. The same estimate holds for condition 4. For conditions 4' and 5' $\theta$ should be substituted by $\gamma \theta$. The probability $p$ that none of $M$ random vectors violates condition 4 can be estimated as $p>1-M\psi_1$ (the same is true for conditions 5, 4' and 5').

Stochastic modelling of the relevant stimuli content $\mathcal{Y}$ could be more sophisticated. First of all, vectors $\boldsymbol{x} \in \mathcal{Y}$ can have some intrinsic similarity or correlations (form a cluster). Nevertheless, analysis of the relevant stimuli as an i.i.d. sample could be also reasonable:  this assumption means that the association will appear from experience and not from the intrinsic correlations between stimuli. Let $\mathcal{Y}$ be an i.i.d. sample from the uniform distribution in a unit ball. Then the probability of condition 3 can be evaluated by the same inequality   
 (\ref{ClusterSeparationProb}) as for conditions 4 and 5 (with $h=\frac{1}{|\mathcal{Y}|-1}(\theta_0-\theta)$). 
 
 The qualitative difference in the results appears because the number of points $|\mathcal{Y}|$ is included in the exponent: $\psi_1<\frac{1}{2}\left(1-\left(\frac{\theta_0-\theta}{|\mathcal{Y}|}\right)^2\right)^{n/2}\approx \exp\left(-\frac{1}{2}\left(\frac{\theta_0-\theta}{|\mathcal{Y}|}\right)^2 n\right)$. For a given number of relevant stimuli $|\mathcal{Y}|$, the probability of violation of condition 2 decays with $n$ exponentially, but unlike the set of irrelevant stimuli, the set of relevant stimuli $\mathcal{Y}$ cannot grow fast with dimension. According to simple estimate, $|\mathcal{Y}|$ should grow slower than $\sqrt{n}$ to guarantee a selected border of probability $\psi_1$ for given $\theta$ and $\theta_0$. 
 
 Condition 3 is the separability of $\mathcal{Y}$ from the origin by the neuron with the average weight vector $\bar{\boldsymbol{w}}$. It was discussed in the previous section (and in \cite{TyukinBrain2017} in more detail).
 
 The estimates discussed here are not sharp, and the constants can be, perhaps, improved. The scaling laws should be more robust. Select an upper bound of the probability of mistake. The number of independent irrelevant stimuli ($\mathcal{M}$) can grow exponentially with the neuronal dimension without violation of this bound. Condition 3 gives the linear growth of the  allowed number of uncorrelated (independent) relevant stimuli with $n$, and condition 2 requires the lower order of growth: the allowed number of independent relevant stimuli for associative learning grows as $\sqrt{n}$ for constant values $\theta$ and $\theta_0$. 
 
\begin{figure}[!h]
\centering{
 \includegraphics[width=0.9 \textwidth]{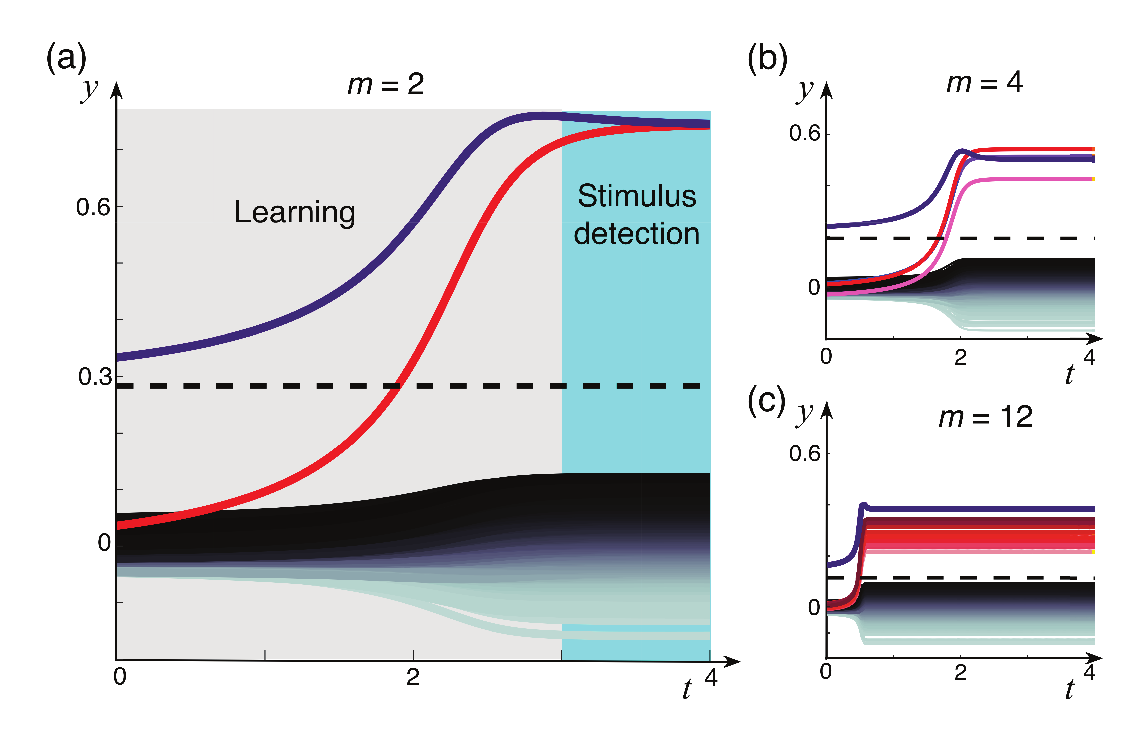}}
\caption{Dynamic memory: Learning new information items by association: a) Example of the dynamic association of a known stimulus ({neuron's response to the known stimulus is shown by} pink curve) and a new one ({neuron's response shown by} red curve). Two relevant stimuli out of 502 are learnt by the neuron. At $t\approx 2$  the orange {curve} crosses the threshold (black dashed line) and {stays above it for $t>2$}. Thus the neuron detects {the corresponding stimulus for $t>2$}; b and c same as in a but for $m = 4$ (b) and $m = 12$ (c), $n=400$, $M = 500$. Gray areas at the bottom -- answers of the neuron to irrelevant stimuli. \label{Fig7}}
\end{figure}

 Figure~\ref{Fig7} shows the results of  numerical experiments  \cite{TyukinBrain2017} of learning by associations  with different numbers of relevant uncorrelated stimuli. In these experiments, $n=400$, $M = 500$.
 
Perhaps, the most important finding is: the experiments of rapid learning by association in hippocampal neurons \cite{IsonQQ2015} with combination of the family member with the Eiffel tower can be reproduced by a simple single neuron learning model. Moreover, this experiment is reproducible with arbitrary (randomly selected) images with high probability, if the neural dimension is high enough. Our numerical experiments showed that `high enough' requirement could be relatively modest and several hundred (or even several dozens) of neurons could be sufficient. 
 
\section{Conclusion}

Ensembles of single neurons can correct mistakes of AI and model neuronal selectivity. They are also used in revealing  neuronal selectivity, fast learning of `concept cells'  by association,  and memory mechanisms. The stochastic separation theorems give the theoretical background of existence and efficiency of `concept cells' and sparse coding in a brain. These theorems provide the next step in a deep generalization of the Gibbs theorems on the equivalence of ensembles after the classical results about concentration of measure (Levy--Milman--Gromov--Talagrand  and other). 

In several words, the stochastic separation theorems state that for an essentially high-dimensional distributions a random point can be separated from a random set by Fisher's linear discriminant with high probability. The number of points in this set can grow exponentially with dimension. Different versions of stochastic separation theorems use  different definitions of `random sets' and `essentially high-dimensional distributions' but the essence of these definitions is simple: sets with very small (vanishing) volume should not have high probability even for large dimension.

The AI correctors (Fig.~\ref{Fig:Corrector}) and concept cells (Fig.~\ref{Fig:Grandmother}) receive signals not directly from the outer world. The distribution of these input signals is neither an inner property of the brain or the AI system, nor a property of the outer world. It is a result of adaptation to real world. During this adaptation, the proper dimensionality reduction should be achieved:
\begin{enumerate}
\item Situations similar in vital aspects should be close after preprocessing;
\item Situations very different in vital aspects should have distant vectors of features;
\item Sets with small volume should not have high probability.
\end{enumerate}

Such dimensionality reduction  is, without a doubt, a `lossy compression'. It can be developed in brain evolution or achieved by machine learning algorithm, but for application of the correctors or for appearance of concept cells, these three conditions should be fulfilled. The third condition can be considered as a byproduct of the first two and the tendency to extend the sensitivity to `vital differences' as much as possible under limited resources. If, after compression, the dimension remains sufficiently high then the single linear elements become useful and blessing of dimensionality appears. 

The  riddle of simplicity is split into the applications of the extended background of statistical physics and into the engineering and reverse engineering technology of lossy compression in preprocessing. We partially understand now the revolution of simplicity (single-cell revolution) in neuroscience and high-dimensional data processing but the design of preprocessing of big data and unravelling the data preprocessing in biological neural nets retain a lot of puzzles. 

One can say that the preprocessing creates the inner world from the outer world, and the concept cells (or small ensembles of cells) translate events in this  inner world into the concepts. This is not the mind yet but an important precursor of the mind.

\section{Acknowledgments}
We are grateful to our coauthors B. Grechuck, E.M. Mirkes, D. Prokhorov, K. Sofeykov, I. Romanenko,  J. Makarova, C. Calvo, A. Golubkov,  and R. Burton. Without our joint work   \cite{GorbanGolubGrechTyu2018,GorbanGrechukTykin2018,GorbanRomBurtTyu2016,GorbTyuProSof2016, TyukinBrain2017, TyuGor2017knowlege}, this review would not have been possible. The work was supported by the Ministry of Education and Science of the Russian Federation (Project No. 14.Y26.31.0022). Work of ANG and IVT was also supported by Innovate UK (Knowledge Transfer Partnership grants KTP009890; KTP010522) and University of Leicester. VAM acknowledges support from the Spanish Ministry of Economy, Industry and Competitiveness (grant FIS2017-82900-P).

 \end{document}